\documentclass[10pt,conference]{IEEEtran}
\usepackage{amsmath,amsfonts}
\usepackage{algorithmic}
\usepackage{graphicx}
\usepackage{textcomp}
\usepackage{xcolor}
\usepackage[hyphens]{url}

\def\BibTeX{{\rm B\kern-.05em{\sc i\kern-.025em b}\kern-.08em
    T\kern-.1667em\lower.7ex\hbox{E}\kern-.125emX}}

\usepackage{color,soul}
\usepackage{enumitem}
\usepackage{multirow, makecell}
\usepackage{pifont}
\newcommand{\cmark}{{\color{olivegreen}\ding{51}}}
\newcommand{\xmark}{{\color{red}\ding{55}}}
\usepackage{booktabs}
\usepackage{fancyhdr}
\usepackage[normalem]{ulem}
\usepackage[hyphens]{url}
\usepackage[keeplastbox]{flushend}
\usepackage[bookmarks=true,breaklinks=true,letterpaper=true,colorlinks,citecolor=blue,linkcolor=blue,urlcolor=blue]{hyperref}

\usepackage{xspace}
\usepackage[capitalize]{cleveref}

\usepackage{tikz}
\usetikzlibrary{patterns}
\usetikzlibrary{backgrounds}

\usepackage{siunitx}

\usepackage{verbatim}
\usepackage{balance}

\usepackage{subcaption}

\newcommand{\scheme}{SGCN\xspace}
\title{\scheme: Exploiting Compressed-Sparse Features in \\ Deep Graph Convolutional Network Accelerators} 

\author{%
\IEEEauthorblockN{%
Mingi Yoo\IEEEauthorrefmark{2}\textsuperscript{,1},
Jaeyong Song\IEEEauthorrefmark{2}\textsuperscript{,1},
Jounghoo Lee\IEEEauthorrefmark{2},
Namhyung Kim\IEEEauthorrefmark{3},
Youngsok Kim\IEEEauthorrefmark{2}\textsuperscript{,*},
and Jinho Lee\IEEEauthorrefmark{4}\textsuperscript{,*}}\vspace{2pt}%
\IEEEauthorblockA{%
\IEEEauthorrefmark{2}\textit{Department of Computer Science, Yonsei University}\\%
\IEEEauthorrefmark{3}\textit{Samsung Electronics}}
\IEEEauthorrefmark{4}\textit{Department of Electrical and Computer Engineering, Seoul National University}\\%
\vspace{2pt}%
\IEEEauthorblockA{%
\{skys7297, jaeyong.song, jounghoolee\}@yonsei.ac.kr, namhyungk11@gmail.com, youngsok@yonsei.ac.kr, leejinho@snu.ac.kr}}
\definecolor{olivegreen}{rgb}{0, 0.6, 0}
\definecolor{grannysmithapple}{rgb}{0.66, 0.89, 0.63}
\definecolor{ceruleanblue}{rgb}{0.16, 0.32, 0.75}
\definecolor{babyblue}{rgb}{0.54, 0.81, 0.94}

\newcommand*\circled[1]{\tikz[baseline=(char.base)]{
            \node[shape=circle,draw,inner sep=0.4pt] (char) {#1};}}
	
\newcommand{\A}{$\mathit{\tilde{A}}$\xspace}
\newcommand{\X}{$\mathit{X}$\xspace}
\newcommand{\Xl}{$\mathit{X^l}$\xspace}
\newcommand{\Xprime}{$\mathit{X}^{\prime}$\xspace}
\newcommand{\XlWl}{$\mathit{X^l}\cdot\mathit{W^l}$\xspace}

\newcommand{\Wl}{$\mathit{W^l}$\xspace}
\newcommand{\AX}{$\mathit{\tilde{A}}\cdot\mathit{X}$\xspace}
\newcommand{\AXl}{$\mathit{\tilde{A}}\cdot\mathit{X^l}$\xspace}

\newcommand{\AXlWl}{$\mathit{\tilde{A}}\cdot\mathit{X^l}\cdot\mathit{W^l}$\xspace}
\newcommand{\Afirst}{$\mathit{(\tilde{A}}\cdot\mathit{X^l})\cdot\mathit{W^l}$\xspace}
\newcommand{\Xfirst}{$\mathit{\tilde{A}}\cdot(\mathit{X^l}\cdot\mathit{W^l})$\xspace}
\newcommand{\format}{BEICSR\xspace} % Bitvector Embedded In-place CSR
	
\newcommand{\JL}[1]{{\color{olivegreen}[\textbf{\sc JLee}: \textit{#1}]}}

\newcommand{\rev}[1]{{\color{olivegreen}#1}}

\definecolor{blue(ncs)}{rgb}{0.0, 0.53, 0.74}
\definecolor{blush}{rgb}{0.87, 0.36, 0.51}
\newcommand{\NK}[1]{{\color{blue(ncs)}[\textbf{\sc NK}: \textit{#1}]}}

\definecolor{carminered}{rgb}{1.0, 0.0, 0.22}

\newcommand{\JS}[1]{{\color{blue(ncs)}[\textbf{\sc JS}: \textit{#1}]}}

\def\final{}   % uncomment this for submission version
\ifdefined\final
\renewcommand{\JL}[1]{}
\renewcommand{\NK}[1]{}
\renewcommand{\JS}[1]{}
\renewcommand{\rev}[1]{#1}
\fi

\newcommand{\lac}{{sparsity-aware cooperation}\xspace}
\newcommand{\Lac}{{Sparsity-aware cooperation}\xspace}
\newcommand{\LAC}{{Sparsity-Aware Cooperation}\xspace}
\begin{document}

\maketitle
\thispagestyle{plain}
\pagestyle{plain}

%%%%%%%%%%%%%%%%%%%%%%%%%%%%%%%%%%%%
% YS: authors' additional information
\begingroup
\renewcommand\thefootnote{1}\footnotetext{Co-first authors.}
\renewcommand\thefootnote{*}\footnotetext{Co-corresponding authors.}
\endgroup
%%%%%%%%%%%%%%%%%%%%%%%%%%%%%%%%%%%%

%%%%%% -- PAPER CONTENT STARTS-- %%%%%%%%

\begin{abstract}
\rev{Graph convolutional networks (GCNs) are becoming increasingly popular as they overcome the limited applicability of prior neural networks.
%A GCN takes as input an arbitrarily structured graph and executes a series of layers which exploit the graph's structure to calculate their output features.
One recent trend in GCNs is the use of deep network architectures.
As opposed to the traditional GCNs, which only span only around two to five layers deep, modern GCNs now incorporate tens to hundreds of layers with the help of residual connections.
From such deep GCNs, we find an important characteristic that they exhibit very high intermediate feature sparsity.
%We observe that with deep layers and residual connections, the number of zeros in the intermediate features sharply increases.
This reveals a new opportunity for accelerators to exploit in GCN executions that was previously not present.}

In this paper, we propose \emph{\scheme}, a fast and energy-efficient GCN accelerator which fully exploits the sparse intermediate features of modern GCNs.
\scheme suggests several techniques to achieve significantly higher performance and energy efficiency than the existing accelerators.
First, \scheme employs a GCN-friendly feature compression format. 
We focus on reducing the off-chip memory traffic, which often is the bottleneck for GCN executions.
Second, we propose microarchitectures for seamlessly handling the compressed feature format.
\rev{Specifically,} we modify the aggregation phase of GCN to process compressed features, and design a combination engine that can output compressed features at no extra memory traffic cost.
Third, to better handle locality in the existence of the varying sparsity, \scheme employs \lac.
\Lac creates a pattern that exhibits multiple reuse windows, such that the cache can capture diverse sizes of working sets and therefore adapt to the varying level of sparsity.
Through a thorough evaluation, we show that \scheme achieves \rev{1.66$\times$ speedup and 44.1\% higher energy efficiency compared to the existing accelerators in geometric mean}.
\end{abstract}

\renewcommand\IEEEkeywordsname{Keywords}

\section{Introduction}

Graph convolutional networks (GCNs) are becoming an increasingly popular type of deep neural networks (DNNs) as they can process highly irregular data~\cite{gcn}.
They have been successfully adopted to achieve high quality in applications such as node/edge classification~\cite{edgepred}, molecular structure analysis~\cite{molecule}, and classic DNN applications including natural language processing~\cite{graphnlp} or scene understanding~\cite{scene}.

Even though GCNs fall into a kind of DNN, it is widely known to exhibit a distinct computational characteristic compared to classic DNNs in that they require dedicated accelerator architectures~\cite{hygcn, dacgcn, bidirectional, engn, awb, igcn, Yoo2021Making, regnn, gcnax}.
The early challenge of GCNs has been to deal with the high sparsity of input graph topology.
It is known that the graph topology data exhibit near-100\% sparsity~\cite{leskovec}.
%Driven by the fact that such sparse graph data induce a lot of random accesses, 
Driven by the high topology sparsity, several proposals have been made to take advantage of the hybrid nature in the execution~\cite{hygcn, dacgcn}, handle load imbalance~\cite{engn, awb}, explore dataflow~\cite{gcnax, Yoo2021Making}, or reorder the topology~\cite{igcn}.

\begin{figure}[t]
    \centering
    \includegraphics[width=0.8\columnwidth]{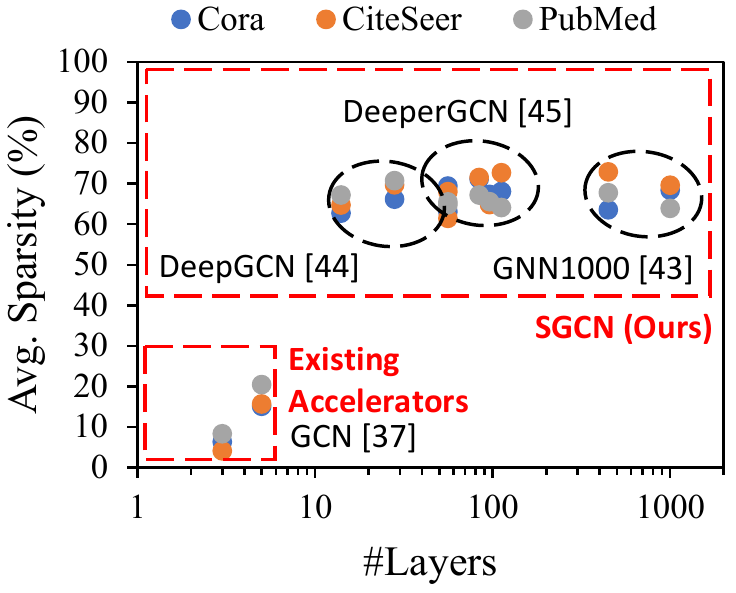}
    \caption{Average sparsity of intermediate features in traditional and modern GCNs with a various number of layers.\JS{120 version fragmented... export it at windows}}
    \label{fig:sparsity}
\end{figure}

%, unlike the highly sparse graph topology data
In contrast to the topology data, the per-vertex feature data has often been treated with dense array representation by existing GCN accelerators because the sparsity of the features is far lower than that of the graph topology~\cite{awb}. %\JL{not exactly true for AWB.. but lets think how we should reformulate this}
Prior efforts targeting the feature accesses were often focused on input features~\cite{awb} or relied on the fact that the feature width shrinks towards the output layer~\cite{awb, bidirectional}.
Those approaches were attractive with traditional GCNs, especially when the number of layers is small ($\leq$ 5) and each layer width is carefully tuned.

However, recent advances in the GCN are leading to different circumstances.
With the introduction of residual connections, modern GCNs~\cite{deepgcn, deepergcn, thousandgcn, gcnii, li2018deeperinsights} now have tens to hundreds, or even a thousand layers, where the feature width remains constant throughout the entire network.

Fortunately, we make an observation that the modern GCNs exhibit much higher intermediate feature sparsity, providing an excellent opportunity for throughput and energy efficiency.
As shown in \cref{fig:sparsity}, the intermediate feature sparsity has been very low in traditional shallow GCNs (up to 20\% sparsity).
However, with deep GCNs with residual connections~\cite{deepgcn, deepergcn, thousandgcn}, the intermediate feature sparsity sharply rises up to 70\% as the network becomes deeper.
When fully exploited, this intermediate feature sparsity can provide higher performance and energy efficiency. 
Considering that the majority of the GCN execution is memory-intensive and shows repetitive random accesses, the expected benefit is significant.

There are several key challenges, however, when it comes to fully \rev{exploiting} the sparsity of the features. 
% #1
First, we need a special format for the GCN intermediate features.
Na\"{i}vely employing existing sparse formats such as compressed sparse row (CSR) for the features may result in lower performance. % and energy efficiency. % as displayed in \cref{fig:moti_vs}.
Even though CSR is considered a de facto standard for sparse data representation, it is not very efficient for the level of sparsity we observe and the dynamic level of sparsity caused from intermediate features.
%Therefore, using CSRs for GCNs easily incurs detrimental effects on both the data size and the performance.
%In addition, provided that GCNs have to access the sparse features randomly, alignment \rev{problems arise and make} it harder to devise the right format.

%CSRs store one entry per each non-zero element, and another entry for its index.
%Therefore, at around 50\% sparsity observed in GCN features, there is no reduction in the number of elements.
%Considering the additional row pointers required for locating rows of varying size, the capacity rises above the original, resulting in poor performance.

% #2 computation is not all. prev scnn is not working  
Second, existing sparse DNN accelerators do not fit well for handling sparse features of GCNs. 
There have been several DNN accelerators, often targeting convolutional neural networks (CNNs)%that utilize the sparsity of the intermediate activation values
~\cite{Parashar2017SCNN, Judd2017Cnvlutin2, Zhou2018CambriconS}. 
However, %directly adopting the techniques is difficult because 
their objectives are usually oriented toward reducing the computational workload.
Such a strategy could be promising to CNNs because they are computationally intensive, and reducing the number of computations (i.e., MACs) easily translates to performance benefits.
However, the computational complexity of GCNs is relatively low, because it usually involves only one MAC operation per each feature element. % loaded from the memory.
In such circumstances, the focus should be on reducing the memory traffic volume, not on reducing the amount of computation.

% %#3 random access and un-alignment
% Lastly, while sparse data representations can easily provide capacity savings, a common problem is the difficulty of supporting random accesses.
% Because the compressed formats yield variable lengths, such formats require additional indirection pointers to denote the starting position of each row in the feature matrix.
% This not only degrades the compression ratio, but also becomes a burden to the on-chip caches because the accesses to the indirection pointers and feature values have conflicting patterns.
% In fact, this also leads to a row-buffer locality problems of the DRAMs, which greatly affects the achieved memory bandwidth.
% When dense representations are used, each access will be several hundreds bytes of regular size, creating a temporarily sequential pattern that is friendly to DRAMs.
% However, with sparse variable length formats, having to alternate between the feature and indirection data increases the risk of row buffer conflicts.

Lastly, the varying level of sparsity makes it difficult to handle locality with tiling techniques.
%The problem becomes more severe with aggressive feature matrix tiling techniques.
Recent work on GCN acceleration~\cite{gcnax, Yoo2021Making} splits the graph topology as well as the feature matrix to reduce the feature working set to fit into cache memory. 
With such techniques, however, the dynamic level of sparsity cannot be estimated at a static time and makes it difficult to determine the right tile size.

% an extra data structure is required to locate the starting positions of each tile, in addition to the rows. 
% In addition, when the tiles are small, the portion of wasted memory bandwidth becomes even larger due to unaligned cache accesses.
% Therefore, for optimized performance, one should also consider the tiling techniques in utilizing the feature sparsity for GCNs.

In this paper, we present \emph{\scheme}, a fast and energy-efficient GCN accelerator which minimizes off-chip DRAM accesses by fully exploiting the intermediate feature sparsity of modern GCNs.
First, we present \emph{Bitmap-index Embedded In-place CSR} (\format) format that is tailored for GCN execution.
Targeting sparsity of roughly around 50\%, we find that using bitmaps as indices result in a good compression rate. 
In addition, by embedding indices in the same row with the content, the accesses can exhibit a much better locality.
Furthermore, \format performs an in-place compression because it suits the parallel writes and random reads from GCN execution.
%With the choice of managing the on-chip memories as caches, the traffic to the off-chip memory is naturally reduced, and the cache can benefit from the reduced working set size. 
%This choice also has the advantage of removing the need for indirection pointers.
%To the best of our knowledge, this is the first work that uses a sparse format for the intermediate features.
Second, we provide microarchitectures and the pipeline structure for executing the sparsity-aware GCNs.
We design a sparse aggregator unit that performs the aggregation phase of GCN from features in a compressed format. %a packed multiply-scattered accumulate.
In addition, we add an in-place compressor with the ReLU unit in the combination phase, such that the compression can be done without having to pay for extra memory access overhead.
Third, we propose \lac for handling the varying working set size due to the dynamic level of sparsity.
We alter the multiple engines to create access patterns containing various-sized working sets. 
As a result, even when the working set size is larger than expected, the cache memory can capture a slightly smaller working set that fits into the cache and reuse some portion of the features.

In summary, this paper makes the following contributions:
\begin{itemize}[noitemsep,nolistsep,leftmargin=*]
    \item We identify the opportunity for utilizing the sparsity of the intermediate features in modern deep GCN execution.
    \item We present \format, a compressed format designed for sparsity-aware aggregation in GCN executions.
    \item We present an \scheme accelerator microarchitecture that benefits from sparse feature data in GCN execution.
    \item We propose \lac to handle the varying working set size induced by dynamic levels of sparsity.
    \item We provide a thorough evaluation for \scheme to show the speedup and energy efficiency gain compared to prior work. 
\end{itemize}

\section{Motivation}
\label{sec:moti}
\subsection{Intermediate Feature Sparsity}
\label{sec:sparsity}

\begin{figure}[t]
    \centering
    \includegraphics[width=\columnwidth]{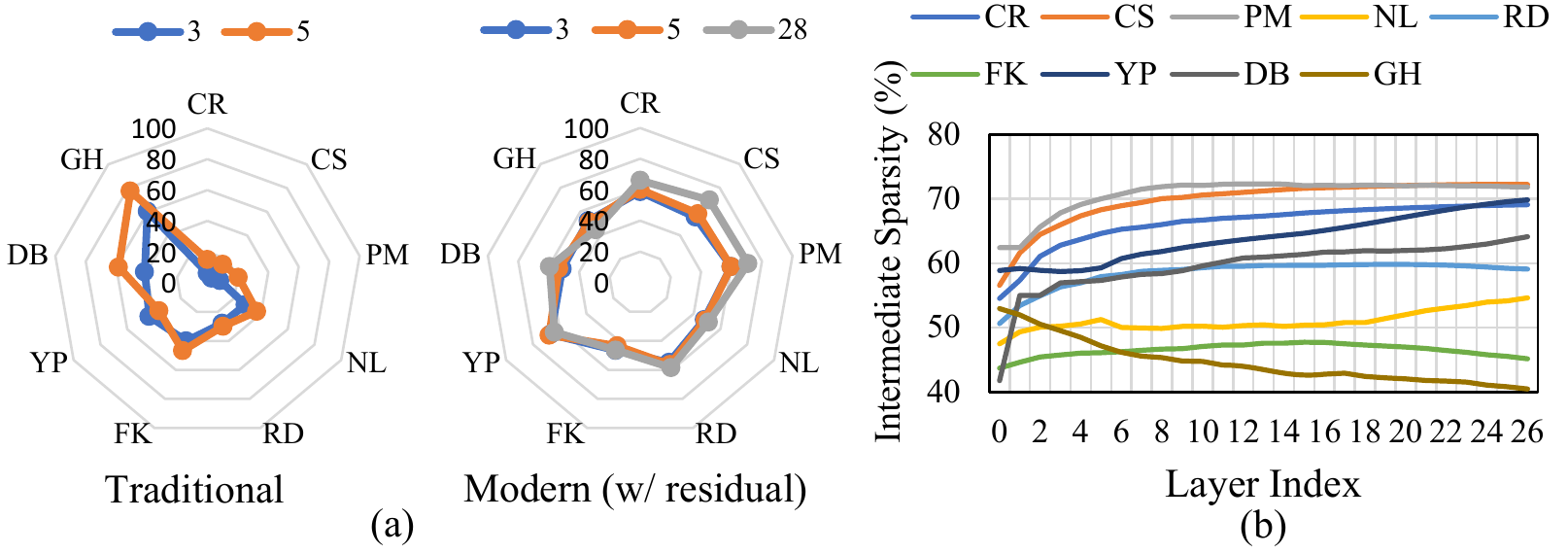}
    \caption{\rev{(a) Effect of residual connection and number of layers on average intermediate feature sparsity and (b) intermediate feature sparsities in layers of modern GCN.}}
    \label{fig:moti_sparsity_analysis}
\end{figure}

% \JL{the purpose of this section is to present a large volume of data, showing that we are not lying}rev{

\rev{In this section, we study the intermediate feature sparsity of deep GCNs.
\cref{fig:moti_sparsity_analysis}a shows the average feature sparsity of traditional GCNs~\cite{gcn} and modern deep GCNs~\cite{deepergcn} whose difference is the existence of the residual connections. 
We chose 3, 5, and 28 layers which are common choices for traditional (3, 5 layers) and modern (28 layers) GCNs.}

As previously shown in \cref{fig:sparsity}, traditional GCNs with less than five layers (\cref{fig:moti_sparsity_analysis}a-Traditional) show low sparsity, which roughly ranges around 5-30\%. 
Note that 28-layer traditional GCN does not converge for all datasets. %In addition, the 28-layer traditional GCN diverged for all datasets.
On the other hand, with modern deep GCNs (\cref{fig:moti_sparsity_analysis}a-Modern), we can make two observations that are distinct from conventional GCNs.
% On the other hand, with modern deep GCNs (\cref{fig:moti_sparsity_analysis} (a)-Modern), we can make two distinct observations from conventional GCNs.
First, the addition of residual connection immediately increases the sparsity to a much higher level~\cite{ressparsity}.
As displayed in \cref{fig:moti_sparsity_analysis}a-Modern, even in shallow conventional GCNs, adding a residual connection yields over 50\% sparsity.
%This is an expected behavior, since the output features are normalized to be averaged at zero. 
When a network is learning well, having around 50\% sparsity is normal behavior, and the low sparsity in the traditional GCNs implies that the network struggles to learn a meaningful set of features.
%Because residual connections are regarded as an essential component that allows using deep layers in GCNs~\cite{gcnii,thousandgcn,deepgcn,deepergcn}, this trend is likely to persist in future GCNs.
Second, deeper networks have more sparsity in general.
Although there are some exceptions, there is a clear trend in \cref{fig:moti_sparsity_analysis}a-Modern that deeper networks exhibit higher sparsity.
In \cref{fig:moti_sparsity_analysis}b, we plot the per-layer sparsity measured from the `28 w/ residual' network.
The results reveal that the sparsity varies per dataset, and is generally sparser towards the output layer. % within a single network.
This aligns with findings from \cite{compressing_dma}, and can be interpreted as that the network is trying to find disentangled representations~\cite{disentangle} for classification tasks.  
%Because the layers closer to the output can draw more information from far neighbors, it is likely that more activations will be disabled. %JL: don't like this. rewrite.

From the two observations, we found that the higher intermediate feature sparsity exists in modern GCNs across datasets, and is likely to exist in future GCNs too.
The goal of \scheme is to utilize this characteristic to design an accelerator with better performance and energy efficiency.

%\JL{should we show the pattern map as well?}

\begin{figure}[t]
    \centering
    \includegraphics[width=\columnwidth]{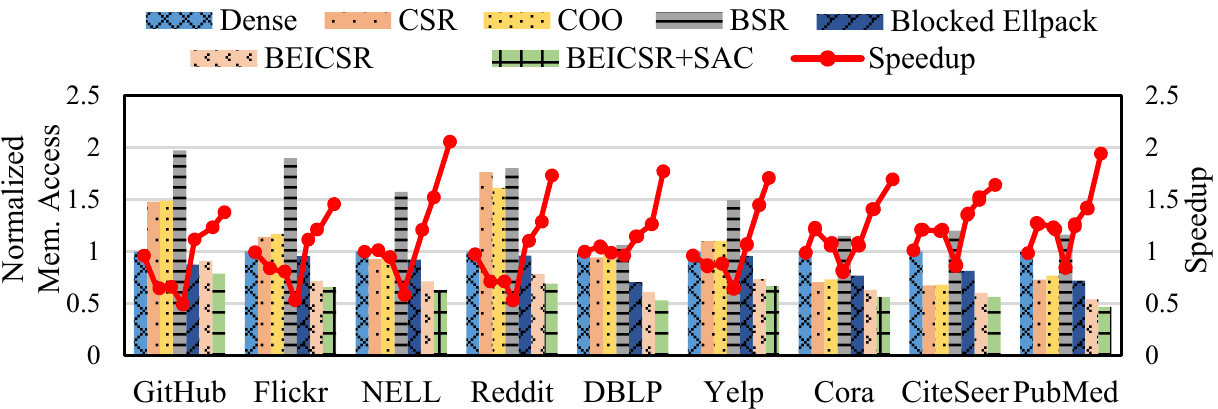}
    \caption{\rev{Comparison of various sparse data representations for GCN intermediate features.}}
    \label{fig:moti_vs}
    \vspace{-2mm}
\end{figure}

\subsection{Need for a New Accelerator Architecture}
In this section, we highlight the need for a new accelerator.
The goal is to exploit the ample amount of intermediate feature sparsity for performance benefits.
\rev{In \cref{fig:moti_vs}, we compare the off-chip memory traffic and the performance of GCN accelerators where dense representation, CSR, COO, BSR, Blocked Ellpack and the proposed sparse format (\format) are used for the intermediate features, respectively.}
We used deep GCNs with 28 layers trained for \rev{nine datasets}, which exhibit around 40 to 70\% sparsity as studied in \cref{sec:sparsity}.
In the plot, we sorted the datasets in the order of increasing average sparsity for clarity.
Please refer to \cref{sec:setup} for the detailed setup.

Despite the large potential from the intermediate features, the results show that na\"{i}vely supporting sparse features on a GCN accelerator results in little or negative speedup.
%In \cref{fig:moti_vs}, we display the result of our motivational experiments on exploiting the sparsity of the features on GCN execution.
The colored bars show the number of off-chip memory accesses caused during the execution, and the curves represent the corresponding speedups.
From the baseline `Dense' which regards the features as a dense matrix, a na\"{i}ve approach would be to adopt a CSR data structure for the features as done for sparse graph topology~\cite{hygcn, awb, gcnax, igcn, Yoo2021Making} or for CNN intermediate features~\cite{scnn, Albericio2016Cnvlutin}.
%\JL{I checked that cambricon does not use CSR. in cambricon-S they say that they put 0 for the neurons}

% However, while using CSR representations could certainly reduce the amount of computation, the na\"{i}ve CSR approach fails to bring much reduction to the memory accesses, due to the overhead the indices required for each non-zero element.
Because the GCN executions are known to be highly memory-intensive, the amount of memory traffic is critical to the performance. 
However, CSR requires an extra element per non-zero value, and fails to provide memory traffic reduction around the provided sparsity range. 
As a result, the na\"{i}ve CSR approach experiences a small or negative speedup from the baseline.
\rev{
The COO format has even more index overheads because it stores both row and column indices for each non-zero element. 
%Because it also requires extra elements per non-zero value.
The problem persists with other sparse formats with block sparsity.
Block Compressed Row (BSR) and Blocked Ellpack are popular blocked sparse representations. 
However, they are beneficial only when there are many empty blocks (e.g., 2$\times$2).   
Because GCN intermediate activations seldom exhibit such patterns, they are not suitable for GCNs.
}
% contains , data that compresses matrix in blocks.
% It is much critical compared to CSR.
% Because It can compress only all of elements in block are zero. 
% For example, in 2$\times$2 block, if one element is non-zero, BSR should store 4 consist block elements.
% Blocked Ellpack is the method that compresses in blocks, but unlike BSR, it only requires the index for the blocks.
% So, the size of Blocked Elpack is much smaller than BSR.
% However, because it share a same weakness to BSR, it still has large memory access than \format. 

% \begin{figure}[t]
%     \centering
%     \includegraphics[width=\columnwidth]{figures/moti_prev.png}
%     \caption{Speedups of previous GCN accelerators which exploited A matrix sparsity.}
%     \label{fig:moti_prev}
% \end{figure}

\rev{
On the other hand, the bars denoted `\format' represent our scheme, which utilizes \format for the feature matrix.
With an efficient compression format designed for the observed level of sparsity, 
our scheme successfully converts the sparsity into a reduction in the memory access count.
Furthermore, the addition of `SAC' (\lac) technique further improves the efficiency of \format.
%Along with the additional \lac technique for increasing the cache efficiency, 
%\scheme provides a great amount of speedup compared to the baseline and the na\"{i}ve CSR approach.
This demonstrates the need for a new accelerator, whose speedup cannot be achieved by existing approaches. 
}

\section{Background}

\subsection{Graph Convolutional Networks}

%\JL{multi layers, eq for residual}

Given a graph topology and a set of incoming features, a conventional graph convolutional network (GCN) produces its outgoing features by performing the operations of its layers on the incoming features.
In a basic GCN, the outgoing features of the $\mathit{l}$-th layer, $\mathit{X^{l+1}}$, is computed as:
\begin{equation}
    \mathit{X^{l+1} = \sigma ( \tilde{A} \cdot X^{l} \cdot W^{l} )}
    \label{eq:gcn}
\end{equation}
where $\mathit{\sigma}$ is a non-linear activation function (e.g., ReLU), often preceded by a normalization function. 
\A{} is the adjacency matrix of the graph, and $\mathit{X^{l}}$ and $\mathit{W^{l}}$ denote the incoming features and weights of the $\mathit{l}$-th layer, respectively.

%\cref{tbl:gcns} depicts the various designs of GCNs.
In the past, it was believed that GCNs deeper than five layers would lead to very low accuracy, a phenomenon is widely known as over-smoothing~\cite{yang2020revisitingoversmoothing, li2018deeperinsights}.
Because of this, the design efforts on GCNs were mostly paid to carefully configuring the feature width of a few layers, usually in a shrinking manner. % as demonstrated in `conventional' models.
In this circumstance, one promising approach was to adjust the order of aggregation and combination~\cite{awb, gcnax, bidirectional}. 
Because changing the order of those two phases did not affect the correctness of ~\cref{eq:gcn}, the combination was done first on a shrinking layer, and aggregation was done first on an expanding layer to reduce the amount of computation and memory accesses~\cite{bidirectional}.
In addition, sometimes specially handling the first layer was found beneficial~\cite{awb}.
Because the first layer's input features (i.e., $X^1$) are not calculated but included in the dataset itself,
those can be super-sparse, especially when one-hot encoding is used.
%Many datasets have one-hot encoded super sparse input features, which is the first incoming feature, 
%In addition, because many datasets have super sparse input features (the first incoming features) often due to its one-hot encoding, specially handling the input features was found to be beneficial for throughput~\cite{awb}.
%

However, as the technology advances, GCNs have evolved to employ residual connections, which alters \cref{eq:gcn} to:
% \begin{equation}
% \mathit{X^{l+1} = \tilde{A} \cdot \sigma(X^{l}) \cdot W^{l} + X^{l}}.
%     \label{eq:res}
% \end{equation}
% \JL{how about this below? the eq above may seem a bit too implementation specific.}
% \begin{align}
% S^{l+1} &= \tilde{A} \cdot X^l \cdot W^{l} + S^{l}, \\
% X^{l} &= \sigma(S^l).
%     \label{eq:res}
% \end{align}
% \JS{How about separating residual, activation, and convolution?}\JL{this looks better}
\begin{align}
S^{l+1} &= S_{res}^{l+1} + S^{l}, \notag \\
S_{res}^{l+1} &= \tilde{A} \cdot X^l \cdot W^{l}, \notag \\
X^{l} &= \sigma(S^l).
    \label{eq:res}
\end{align}

\begin{figure}[t]
    \centering
    \includegraphics[width=\columnwidth]{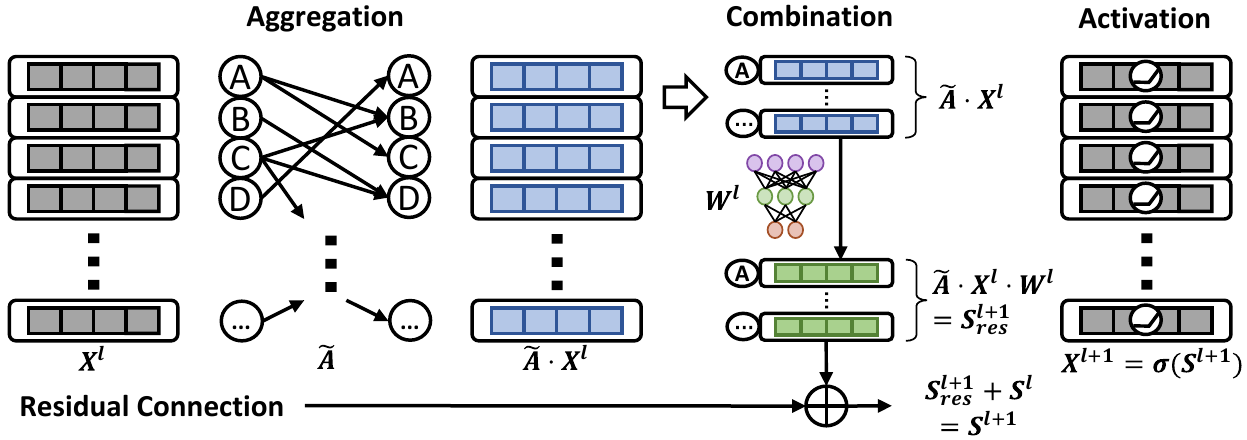}
    \caption{Operations of a modern GCN layer. \vspace{-2mm}}
    \label{fig:background:gcn}
\end{figure}

Fig.~\ref{fig:background:gcn} illustrates the operation.
The calculation of \AXl and its variants are usually called \emph{aggregation}, which does not involve trainable weights.
It is the process of collecting the features from neighboring vertices and aggregating them.
The calculation of \XlWl and its variants are often called \emph{combination} as it computes the next layer's features by linearly combining the incoming features and the weight parameters.
%

%\NKr{accuracy of GCNs? - reviewer A}
With the introduction of residual connections, it was found that GCNs can be as deep as convolutional neural networks, and tens~\cite{gcnii, deepgcn, deepergcn} to even a thousand layers~\cite{thousandgcn} are being used for state-of-the-art GCNs.
With deep layers, these designs now have uniform feature widths, which invalidates the phase ordering technique.
Furthermore, the sparsity of the input features now does not have much impact on the total execution time, as the speedup would be amortized over the tens of layers.
Fortunately, this use of residual connections dramatically increases the sparsity of the intermediate features, which enables another opportunity for performance improvement of GCNs. 
% \JS{Maybe we need to discuss about the Res+, Res, and Dense deepgcn residual connection.}

%\JL{I removed: 1. the claim about focusing on aggregation. not needed any more. 2. 50\% sparsity. to be moved to motivation} 

% Among the two phases, the optimization efforts into GCN acceleration are often weighted toward
% the aggregation part for the following reasons.
% First, the execution bottleneck is usually at the aggregation, as analyzed in previous work
% ~\cite{hygcn}, due to the aggregation’s random-access nature, making its execution memory-bound and hard to benefit from caching.
% Second, the combination is not much different from conventional DNNs.
% Thus, the acceleration of the combination part can typically adopt the architecture of the existing DNN accelerators~\cite{tpu, eyeriss, Albericio2016Cnvlutin}.
% Following these analyses, we also focus on improving the performance of the aggregation.

% Most of the real-world graph datasets are known to be extremely sparse, and this leads to the sparsity of \A being almost 100\%~\cite{leskovec}, which becomes a reason for designing dedicated GCN accelerators. 
% On the other hand, the feature matrix \X exhibits around 50\% sparsity from the common use of ReLU as the activation function.
% Because most of the memory accesses in the aggregation part come from randomly accessed rows of the feature matrix, exploiting the sparsity has a huge potential for the performance of the GCN execution.

\subsection{Hardware Acceleration of GCNs}
\label{sec:acc}
Often, the key to the GCN accelerators is supporting the hybrid nature of aggregation and combined phases.
Some architectures implement shared units that support varying datapath for both phases~\cite{awb, gcnax, engn}, and others implement separate units and pipeline two phases~\cite{hygcn, regnn, igcn, Yoo2021Making}. 

Table~\ref{tab:summary} summarizes the characteristics of the existing GCN accelerators.
GCN accelerators can be further classified into two types depending on which of the aggregation and combination phases get performed first.
\emph{Aggregation-first} accelerators perform the aggregation phase followed by the combination phase.
That is, an aggregation-first accelerator first computes \AXl followed by \Afirst.
\emph{Combination-first} accelerators, on the other hand, compute \XlWl followed by \Xfirst.
The advantage of each path lies in whether the width of the features 
increases or decreases after the combination phase, and some approaches suggest utilizing the differences~\cite{gcnax, bidirectional}. 
However, with the introduction of deep GCNs~\cite{gcnii, deepgcn, deepergcn, thousandgcn}, the intermediate feature widths usually remain the same throughout the network, and the differences disappear.

\begin{table}[b]
    % \scriptsize
    \footnotesize
    % \small
    \centering
    \caption{Comparison of GCN Accelerators}
    \label{tab:summary}
    \begin{tabular}{lcccccccc}
        \toprule
        \multirow{2}{*}{\textbf{Accelerator}} &
        \multirow{2}{*}{\makecell{\textbf{Compressed}\\\textbf{Feature?}}}  &
        \multirow{2}{*}{\makecell{\textbf{Target}\\\textbf{Layers}}}  &
        \multirow{2}{*}{\makecell{\textbf{Residual}}}  &
        \multirow{2}{*}{\makecell{\textbf{Execution}\\\textbf{Order}}} \\
        % \cmidrule(lr){2-3}
        % \cmidrule(lr){4-5}
        % & \multicolumn{2}{c}{\A{}} & \multicolumn{2}{c}{\X{}} & \\
        \\
         \midrule
        AWB-GCN~\cite{awb} &\xmark{} & 2 & \xmark{} &  Comb. first \\
        EnGN~\cite{engn} &  \xmark{} & 2 & \xmark{} & Comb. first \\
        HyGCN~\cite{hygcn} &  \xmark{} & 1-2 & \xmark{} &  Aggr. first \\
        Bidirectional~\cite{bidirectional} &  \xmark{} & 3 & \xmark{} &  Both \\
        GCNAX~\cite{gcnax} &  \xmark{} & 2 & \xmark{} &  Both \\
        % Yoo et al.~\cite{Yoo2021Making}  & \xmark{} &N/A & \xmark{} & Comb. first \\
        I-GCN~\cite{igcn} & \xmark{} & 2 & \xmark{} &  Comb. first \\
        % ReGNN~\cite{regnn} & \xmark{} & 2 & \xmark{} & Aggr. first \\
        \midrule
        \textbf{\scheme (Ours)} & \textbf{\cmark{}\makecell{\format \\ (\S \ref{sec:format})}} & \textbf{$>$5} & \textbf{\cmark{}} & \textbf{Aggr. first} \\
        \bottomrule
    \end{tabular}
\end{table}

\begin{figure}[t]
    \centering
    \includegraphics[width=\columnwidth]{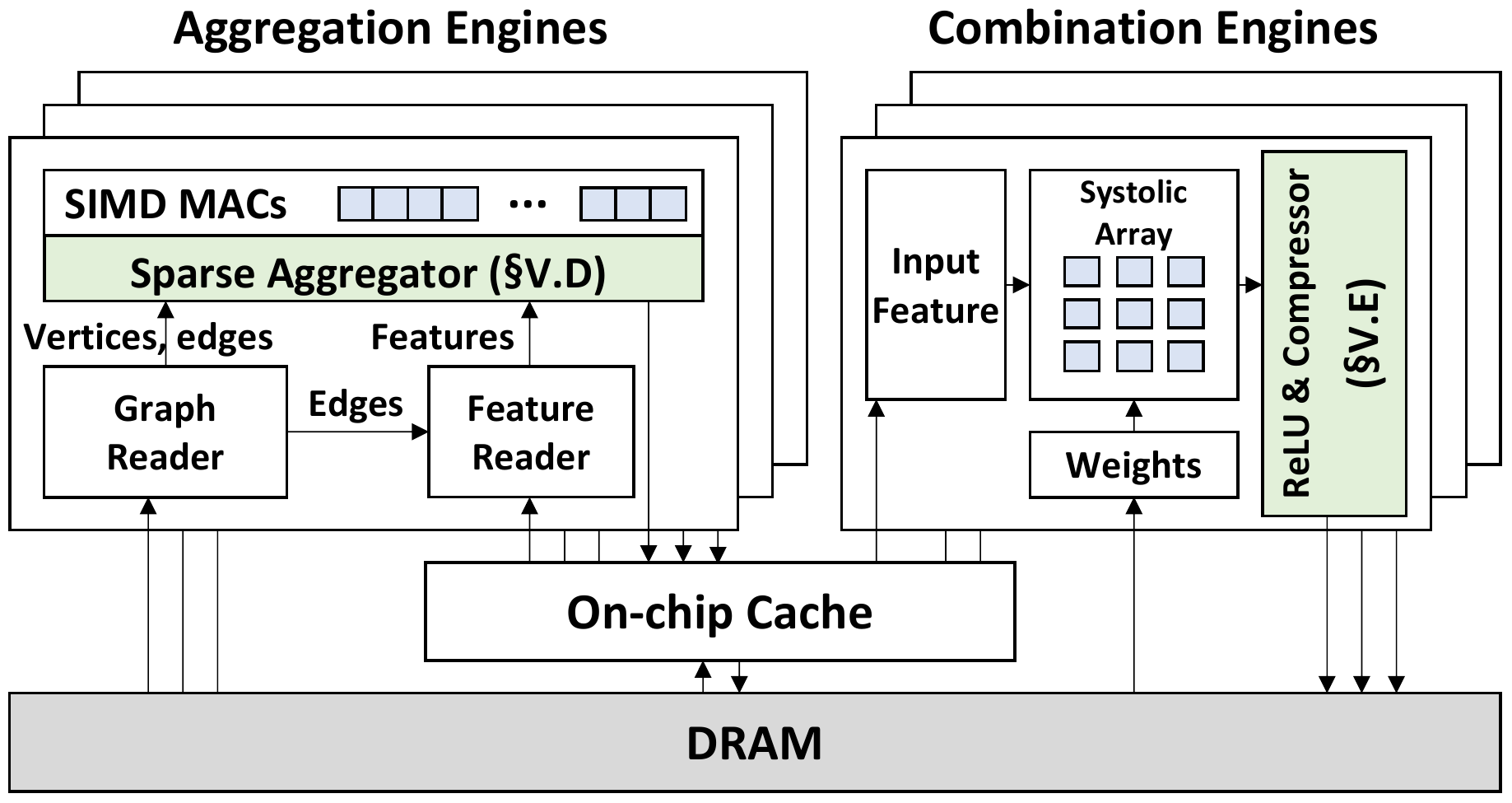}
    \caption{Our GCN accelerator architecture exploits the high sparsity of graphs. Green modules are added for \scheme.}
    \label{fig:baseline:arch}
    \vspace{-3mm}
\end{figure}

Fig.~\ref{fig:baseline:arch} shows the baseline GCN accelerator architecture, similar to prior art~\cite{hygcn, regnn, Yoo2021Making}, where the green shaded units have been added for \scheme.
The aggregation unit uses SIMD MAC cores to process the accumulation of features from multiple vertices.
The topology matrix \A is assumed to be in a CSR format to employ the high sparsity. 
Similar to graph processing accelerators~\cite{graphicionado, ozdal, extrav}, a graph reader reads the indices vertex indices and the corresponding edges.
From the edge information, the feature reader fetches the feature vectors of the edge destinations. % for the SIMD cores. 
Together, these modules feed the SIMD cores to continuously process the aggregation without being stalled.
% Together, these modules feed the SIMD cores to process the aggregation without being stalled continuously.
Each module has a small buffer to temporarily store prefetched values to avoid stalls from upstream backpressure.
%\JL{add edgepref $\rightarrow$ featureread in fig}

As accesses to the feature vectors from neighboring vertices exhibit a highly randomized pattern over a wide range of data, a sizable on-chip memory is used as a global cache resembling a last-level cache in modern CPUs.
However, the working set size often far exceeds the capacity of the global cache, prohibiting an efficient use of the inherent locality in the feature vector accesses, leaving the aggregation phase highly memory intensive.

The combination engine contains a systolic array for matrix multiplications at its core, similar to conventional DNN accelerators~\cite{tpu, maeri}. 
The input feature and weight buffers provide input matrices, $X$ and $W$, respectively, to the systolic array. 
The output is written back to the off-chip DRAM, usually becoming the input to the next layer.

\section{Design Goals}

\begin{figure*}[t]
\centering
\begin{subfigure}[t]{.37\textwidth}
\hspace{-2mm}
% \vspace{-5mm}
% \centering
\includegraphics[height=1.95cm]{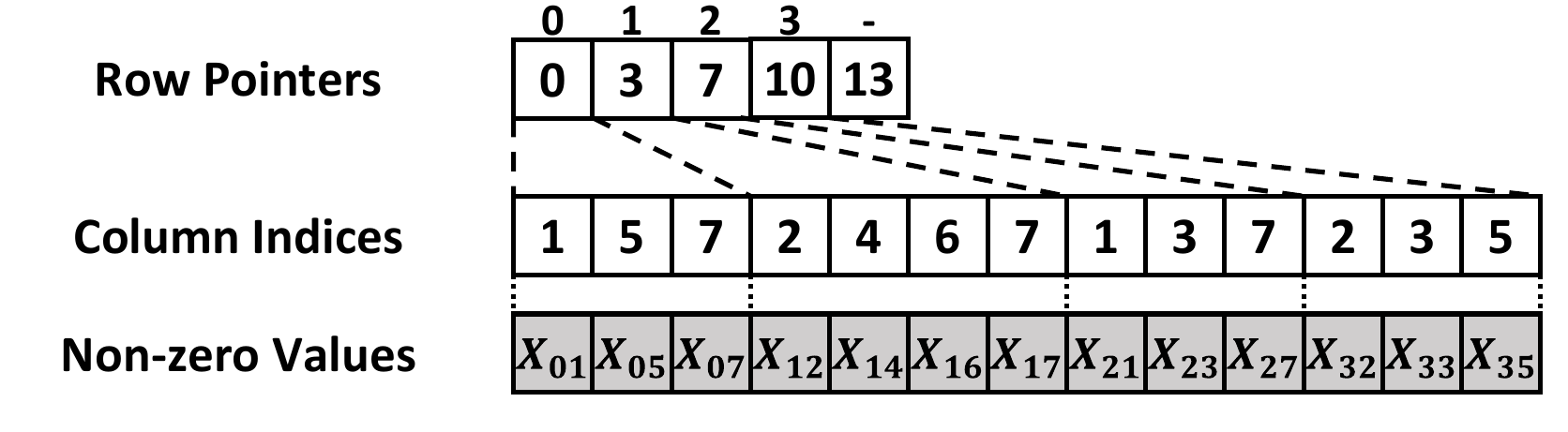}
\caption{CSR format.}
\label{fig:csr}
\end{subfigure}
\begin{subfigure}[t]{.27\textwidth}
% \vspace{-5mm}
\centering
\includegraphics[height=2.1cm, width = 4.85cm]{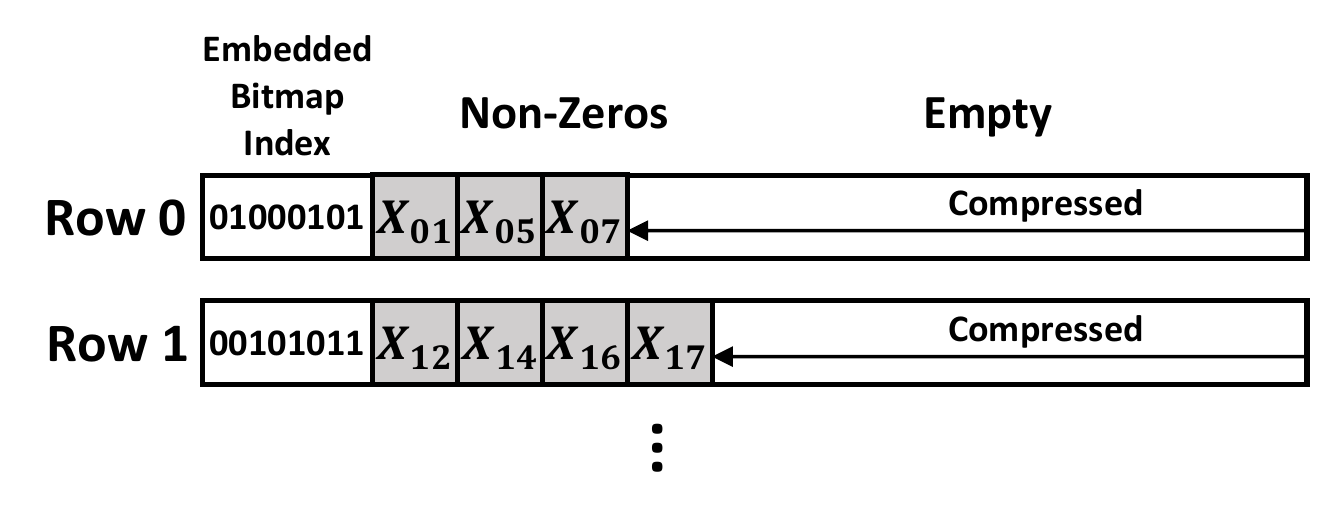}
\caption{\format format.}
\label{fig:bicsr}
\end{subfigure}
\begin{subfigure}[t]{.32\textwidth}
% \vspace{-5mm}
\centering
\includegraphics[height=1.95cm]{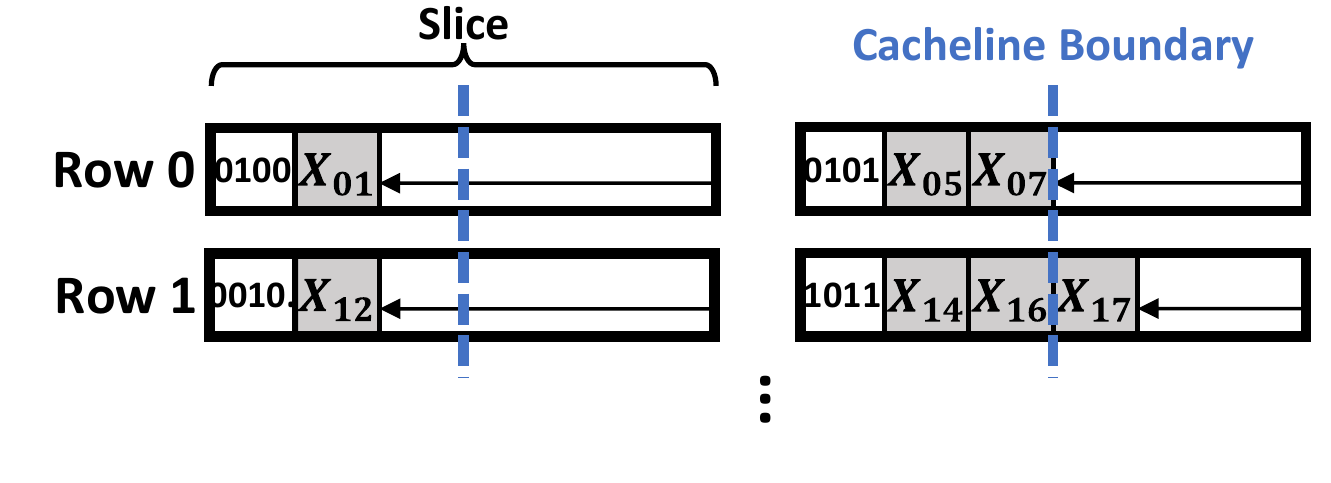}
\caption{\format with feature tiling support.}
\label{fig:bicsr_slice}
\end{subfigure}
    \caption{Compression formats of the feature matrix.}
    \label{fig:format}
    \vspace{-3mm}
\end{figure*}

\label{sec:goals}
\JL{title:Challenges and design goals?}
From the motivation, our objective is to \textbf{devise a sparse format} for the features and \textbf{design an accelerator microarchitecture} that efficiently process them.
For the objective, the following design goals were set in deriving the schemes for \scheme.

First, the primary target should be the memory access reduction, not the computation or the capacity.
The primary bottleneck of GCN execution is known to be the aggregation phase, which is extremely memory intensive~\cite{regnn, igcn}. 
This indicates that solely optimizing the amount of computation without memory traffic reduction (e.g., zero skipping) would not be an optimal choice.

Second, the format and the resulting access pattern should be cache- and \rev{DRAM}-friendly.
In contrast to CNNs where the feature accesses can be done sequentially, GCN aggregation features incur random accesses of small per-vertex features (i.e., a few cachelines).
When dealing with sparse formats, special care must be taken that each access is aligned to the cacheline and/or the DRAM burst length to optimize off-chip traffic.
Furthermore, because the intermediate features are dynamically created as a result of the previous layer, the format should allow parallel writes as well as reads. 
Especially with variable length rows as a result of compression, it is easy for the access pattern to exhibit many unaligned accesses or bank conflicting accesses.
Such a problem appears more often for modern memory systems such as HBM that do not support multi-rank channels.
Thus, the compression format and its execution should be aware of the memory subsystem and exploit it.

Third, the execution should embrace existing GCN accelerators.
For example, existing accelerators employ overlapping between phases~\cite{hygcn, gcnax} or topology/feature tiling~\cite{gcnax, Yoo2021Making}.
Na\"{i}vely applying compressed format and execution flow could potentially disable such techniques, especially with variable sparsity levels that cannot be estimated statically.
With the sparsity-aware GCN execution, such techniques must be embraced to achieve maximum performance.

\section{\scheme Architecture}

% \begin{figure}[t]
%     \centering
%     \includegraphics[width=\columnwidth]{figures/cache_need.png}
%     \caption{Need for cache by comparing ours and others}
%     \label{fig:cacheneed}
% \end{figure}

% architecture first, or last?
% i.e. top-down or bottom-up?

% cambriconx : weight only

% cnvlutin: activation only. ZFNAf encodes neurons as (value,offset) pairs in groups called bricks. Bricks are stored starting
% at the position their first neuron.

% cnvlutin2: for activation, 
% 1. 0/1 compress or not
% 2. bitvec + store zeros too
% 3. bitvec + store non-zeros + indirection

% scnn: csr

\subsection{Compression Format}
\label{sec:format}
\JL{memo:if not already we should say that in-place DS is needed because this is 1. dynamic sparsity and 2. random access somewhere..}

\JL{are we saying anywhere the name \format???}

The ultimate purpose of compression from \scheme is to reduce off-chip memory access. 
This is obviously sensitive to the compression format which we employ for the feature matrix.
Compared to the conventional sparse DNN accelerators~\cite{Parashar2017SCNN,Judd2017Cnvlutin2}, the difference is that the feature matrix of GCNs exhibit a highly randomized pattern due to the extremely sparse nature of the topology matrix \A.
For each random access, a portion of a vertex feature array is accessed, which usually spans a few cachelines.

A na\"{i}ve choice for dealing with sparse data is a compressed sparse row (CSR).
However, applying a na\"{i}ve CSR will not result in off-chip traffic reduction.
With CSR, every non-zero element requires two values: one for the column index, and the other for the value itself. 
%Considering that there is a row pointer for every row, 
Thus, at sparsity below 50\%, there is only capacity overhead instead of reduction, as exemplified in \cref{sec:moti}.
%An optimization to this would be using reduced bitwidth for the column indices. 
%However, it would limit the width of the feature matrix, and still cause a significant overhead of ~10 bits per non-zero element. %moved to discussion

To tackle the above issues, we propose \format (Bitmap-index Embedded In-place CSR) format for the intermediate GCN features.
\cref{fig:format} illustrates the conventional CSR format (6a) and the compression format proposed for \scheme (6b) against the same set of data.
We make the following design choices for the format:

\textbf{Embedded Bitmap Index.} 
% #1. Bitvector
Instead of column indices, we use \emph{bitmap indices} for each vertex.  
For example, if an example array of size four is $(0, 0.3, 0.5, 0)$, we place $0110'b$ followed by the non-zero array $(0.3, 0.5)$.
Assuming the feature vector has elements with 50\% sparsity, and each element occupies 32 bits, we can calculate the overhead of the bit vector index.
when the feature vector width is $n$,
%Assuming the feature vector has $n$ 32-bit elements with 50\% sparsity, %, and each element occupies 32 bits, we can calculate the overhead of bit vector index.
%When the feature vector width is $n$, 
the bitmap index size is $n$ bits, and the non-zero feature data size is $32n \times 0.5 = 16n$. 
Therefore, it leads to a total index overhead of only 6.25\% in that case ($n / 16n$),
which would be much lower than that of the na\"{i}ve CSRs that require one integer index per non-zero element.
On the other hand, if we use CSR for the same 50\% sparsity, we have $32n \times 0.5$ overhead for the column indices plus the row pointers to indicate the starting location of the arrays.
Combined with the non-zero feature data size, it would result in an increased size instead of compression. \JL{rethink the comparison}

%#2. Embedding.
In addition, we pose to \emph{embed the bitmap index} at the head of the same array which stores the non-zero values.
In compressed formats, indices are commonly placed in an independent array. %Although bit vector has less overhead, naively allocation with an independent array for bit vector indices 
However, \rev{such a choice} would result in a poor memory access pattern.
When a bit vector index for a row is accessed, the surrounding cache line containing indices for other rows is always accessed together in single memory access.
Unless reused before eviction, they are considered overhead.
Observing the access pattern, the accesses to the bit vector index are almost always followed by the non-zero values.
The only exception is when the bit vector index contains all zeros.  (i.e., no non-zero element in the row).
However, considering that the sparsity is around 50\% and each element has little dependence on the other, such an occasion is unlikely.
Therefore, the choice of embedding the bitmap index in the same array with the non-zero values yields a better memory access pattern.
\rev{Note that there exist some formats using bitmap indices~\cite{bitmap1,bitmap2}, but they rely on entirely empty rows~\cite{bitmap1} or blocks~\cite{bitmap2} and hence are inappropriate for GCN features. }
% \MG {I added the explanation but need to cleaning} \MG {dose the overhead realy need? because we use BEICSR with feature tiling support, I think the overhead is not important.}
% \MG {I combined Embedded Bitmap Indice section}
% Assuming the sparsity of the feature matrix of about 50\%, the overhead of such a bit vector would be approximately two bits per non-zero element, which is about 6.25\% of the whole feature data. \NK{how can we define overhead of compressed data?} \MG{assume each of elements is 4Byte(32bit) float. in the example, the number of non-zero elements are 2(=64bit). because feature array size is 4, extra 4 bits is needed for bitmap. so, the bitmap overhead is 4/64 = 0.0625 = 6.25\% }\JL{@MG: would you put a very short calculation on the body?}\JL{compare with CSR @ 50\%}

\textbf{In-place Compression.} A common problem with using compressed formats for performance is the variable lengths.
Not only this would necessitate an indirection array similar to a row pointer, but also result in frequent misaligned memory accesses that lead to traffic volume overhead.
To store the offsets for each row, some indirection array equivalent to a row pointer would be needed. 
Instead, we compress the data row-by-row and store each row at the same reserved place in the memory as if the row was left uncompressed.
Even though this would give no benefit to the memory capacity, 
we argue that \rev{such a choice} is almost necessary for the following reasons. %\JL{is it too strong?}
First, it would provide reduced off-chip traffic aligned to the cacheline boundaries. 
The alignment is especially important, since the access granularity only spans a few cachelines.  
Much of the space reserved for a row would be empty, and therefore not accessed from the memory.
Furthermore, the beginning of each row can be cacheline-aligned, such that the loss from misalignment can be minimized.
Second, it allows parallel writes at the output of each layer. 
Naively using a variable length storage format requires serialization, 
because we do not know the size of the compressed feature array from each vertex.
Such a choice would incur an intolerable overhead to the execution time.
Lastly, because the sizes of the rows are uniform, there is no need for an indirection array. 
Locating the compressed data just involves a multiplication with the vertex id, which eliminates the need for accessing indirection arrays. 
This scheme reduces the memory access count and contributes towards row-buffer locality, and achieves better memory bandwidth utilization.

\JL{I need to say burst length (32B) instead of cacheline size (64B) all thruout the paper}

% \textbf{Embedded Bitmap Indices.}\NK{merge section with bit vecor index}
% Unlike common CSRs, we embed the bit vector index at the head of the same array which stores the non-zero values.
% Employing a data structure similar to a CSR, one could naively allocate an independent array of the bit vector indices.
% However, this scheme would result in a poor memory access pattern.
% When a bit vector index for a row is accessed, the surrounding cacheline is accessed together in single memory access.
% Unless reused before eviction, they are considered overhead.

% Observing the access pattern, the accesses to the bit vector index is almost always followed by the non-zero values.
% The only exception is when the bit vector index contains all zeros (i.e., no non-zero element in the row).
% However, considering that the sparsity is around 50\% and each element has little dependence on the other, such an occasion is unlikely.
% Therefore, putting the bit vectors and the non-zero together yields a better memory access pattern.

\subsection{Supporting Feature Matrix Slicing}
\label{sec:support}

Many modern GCN accelerators propose slicing the feature matrix for better data reuse~\cite{snf, Yoo2021Making, gcnax}.
When compressing the entire row together, accessing a slice of the feature matrix involves the following sequence: 1) Read the bit vector index to find the range of non-zero values corresponding to the slice, 2) read multiple cachelines that contain non-zero values from the slice, and \rev{3)} perform aggregation.

The pitfall from the above sequence is the overhead of unaligned accesses. % becomes non-negligible.
For example, when there are 16 non-zero values (64 bytes) in a slice, unaligned access would almost always require two 64B cachelines, which neutralizes the benefit from the sparsity.

Therefore, we employ \emph{sliced \format} as depicted in \cref{fig:format} (c).
Instead of a single set of bit vector indices for the entire row, we partition the bit vector and embed it in the head of each corresponding unit slice. 
Then we align the slices to the burst boundaries.
Using the in-place compression from \cref{sec:format} for each slice, we allocate the memory space that can hold the maximum number of non-zeros within them (i.e., a dense slice).
With the right choice of the unit slice size $C$, the wasted amount of memory access can be minimized because the number of non-zero elements has a small variance and there are only a few outliers.
When a larger slice size is desired, we can simply combine multiple unit slices to form a large logical slice, which incurs almost no difference compared to having a large slice in the first place.
In the default design of \scheme, we empirically set $C=96$, which would occupy 384 bytes in single precision features, and with around 50\% sparsity it fits into two or three cachelines along with the embedded bitmap indices.

\begin{figure}
 \centering
  \includegraphics[width=0.5\textwidth]{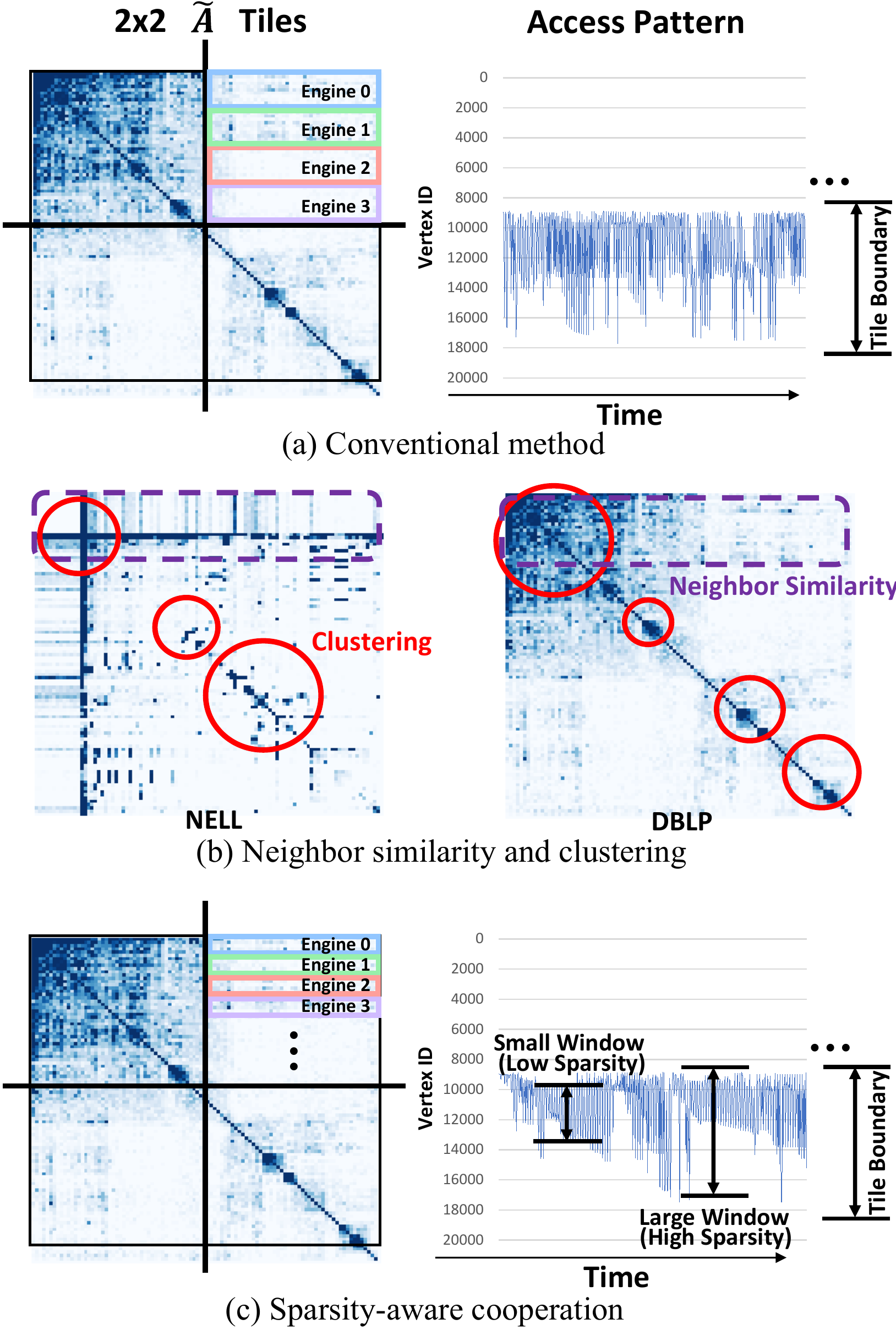}
 % \vspace{-3mm}
  \caption{(a) Conventional method, (b) density heatmap from adjacency matrices of the NELL and DBLP graph, and (c) \lac assuming 2$\times$2 tiling. 
  %With conventional method, the working set size is determined statically. %However, with \lac, the cache can adapt to different working set sizes induced by varying sparsity.
\JL{A -> 2x2 A tiles}\JL{explain how to read it in the body}  }
  %\NK{what is this figure about? engine->tile. window=working set size before compression? cache can capture large window @b, how about large window @a?} \MG {because of different x sparisity, window size is dynamically changed. but (a), it accesses almost always large window}
 % \vspace{0 mm}
  \label{fig:lac}
   \end{figure}   

% \begin{figure}
% \centering
%  \includegraphics[width=0.9\columnwidth]{figures/adjacency.pdf}
%  \caption{Density heatmap from adjacency matrices of the NELL and DBLP graph dataset. A darker color indicates a denser region.}
%  \label{fig:adjacency}
%   \end{figure}

\subsection{\LAC for Graph Topology Tiling}
\label{sec:lac}
Another promising technique for GCN execution is graph topology tiling~\cite{gridgraph, gcnax, awb}.
Partitioning the adjacency matrix into multiple tiles helps reduce the working set size of the intermediate features such that it fits into the cache size.
Existing approaches usually find the optimal tile size based on an off-line analysis, often by statically calculating the working set size~\cite{Yoo2021Making, gcnax}.
However, with \scheme, the level of sparsity varies dynamically depending on each vertex, dataset, and layer.
Because of this, estimating the optimal tile size is very difficult, and a working set size exceeding the cache capacity results in significant performance degradation. %, and using a single tile size for entire network does not yield good performance.

\cref{fig:lac}a shows how conventional graph tiling works on an example of $2\times2$ tiling.
On the right side, it shows vertex IDs of the features that are accessed for the first 300 reads.
When a graph is partitioned into tiles, the working set size is confined to the features of the vertices included in a tile, which can fit into the cache.
However, when the sparsity of the features is lower than expected, the effective working set size increases because there are more non-zeros per vertex.
This has the risk of exceeding the cache capacity, and the performance would quickly drop due to the thrashing pattern~\cite{qureshi_adaptive_insertion,rrip,ship}.

To address this issue, we propose \emph{\lac}, a method that alters the access pattern such that there are variously sized working sets that can be captured by the caches.
It is known that graphs often form community clusters~\cite{girvan2002community}, and there exists a neighbor similarity between adjacent vertices~\cite{bv}.
For example, the example density map from NELL~\cite{nell} and DBLP~\cite{dblp} displayed in \cref{fig:lac}b shows that adjacent rows tend to \rev{exhibit the same} patterns, and have strong clustering around the diagonals. %\JL{fix to uppercase NELL}
%\NK{Fig.7 <-> Fig.8 change order?} \MG{yes. I just changed them}
%\lac takes advantage of this and let alter the data access pattern such that the locality gets less affected by the explicit tile size.

% \cref{fig:lac} illustrates how \lac works.
% In a conventional scheme, multiple tiles are assigned different set of vertices within a tile.
% To keep the accesses within a cache, the tile size is usually set such that the corresponding feature access window sizei fits the cache size.
% However, if miscalculated, the cache efficiency quickly drops as the order of accesses is similar to a thrashing pattern~\cite{qureshi_adaptive_insertion,rrip,ship}.

\Lac takes advantage of this, and each engine accesses a small strip of vertices in an interleaved manner as illustrated in \cref{fig:lac}c. 
For the height of strips, we empirically use 32.
Because of the neighbor similarity and clustering, 
the access distances tend to be shorter, and there are multiple working sets with diverse sizes that can be captured. %the effective access window gradually shift together.
As a result, when the sparsity level is high, the cache will capture the larger working set window (denoted as `large window'), and when the sparsity is low, the cache can capture a smaller working set window to avoid thrashing (denoted as `small window').

  \begin{figure}[t]
    \centering
    \includegraphics[width=0.95\columnwidth]{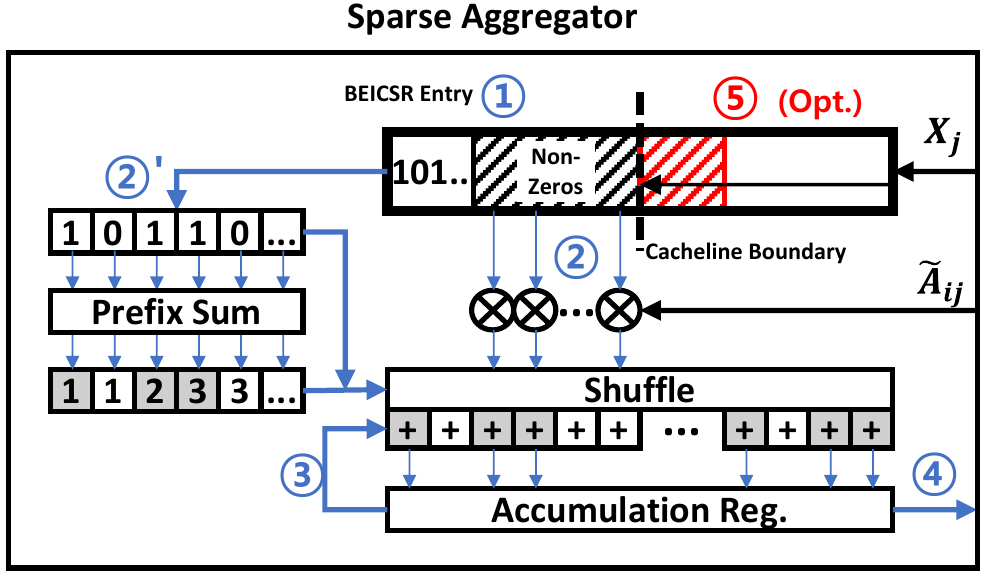}
    \caption{\scheme sparse aggregator unit microarchitecture.}
    \label{fig:arch}
\end{figure}

% \JL{here}
\subsection{Sparse Aggregation}
\label{sec:spagg}
\cref{fig:arch} reveals the microarchitecture of the sparse aggregator unit in \scheme and its execution procedure.
Each sparse aggregator of \scheme has 16 multipliers, which can process a single cache line worth of data together.
% \NK{how many element in a cacheline? 16 * 4B/elem + bitmap > 64B?} \MG {Less elements are in a cacheline. if bitmap size is 32bit, only 15 elements can be in a cacheline. (it's not a quite overhead)}
\circled{1} When a row of the feature matrix \X is selected, its first 64 bytes are fetched to the aggregation engine, where its head contains the bitmaps and the rest contains the non-zero values.
\circled{2} The non-zero values are multiplied with the corresponding edge weight of \A broadcasted each multiplier. 
$\circled{2}^\prime$ In parallel, the bitmap is processed by a parallel prefix sum unit to convert the 1's in the bitmap to a reversed\NK{?}\JL{lets fix}\JL{I think it's reversed..} index to the non-zero values.
\circled{3} The bitmap and the reversed indices are sent to the accumulator. 
If the bitmap value is 1, the accumulators at the corresponding positions load the multiplier outputs and add them to the current value.
\circled{4} When the accumulation for a single vertex is complete, it is sent to the combination engine for performing combination, ReLU, and compression.
\circled{5} (Optional) When there are still non-zeros remaining in the next cacheline (identified by the prefix sum result), the next 64 bits are fetched to perform \circled{1}-\circled{4} again.

\begin{figure}[t]
    \centering
    \includegraphics[width=\columnwidth]{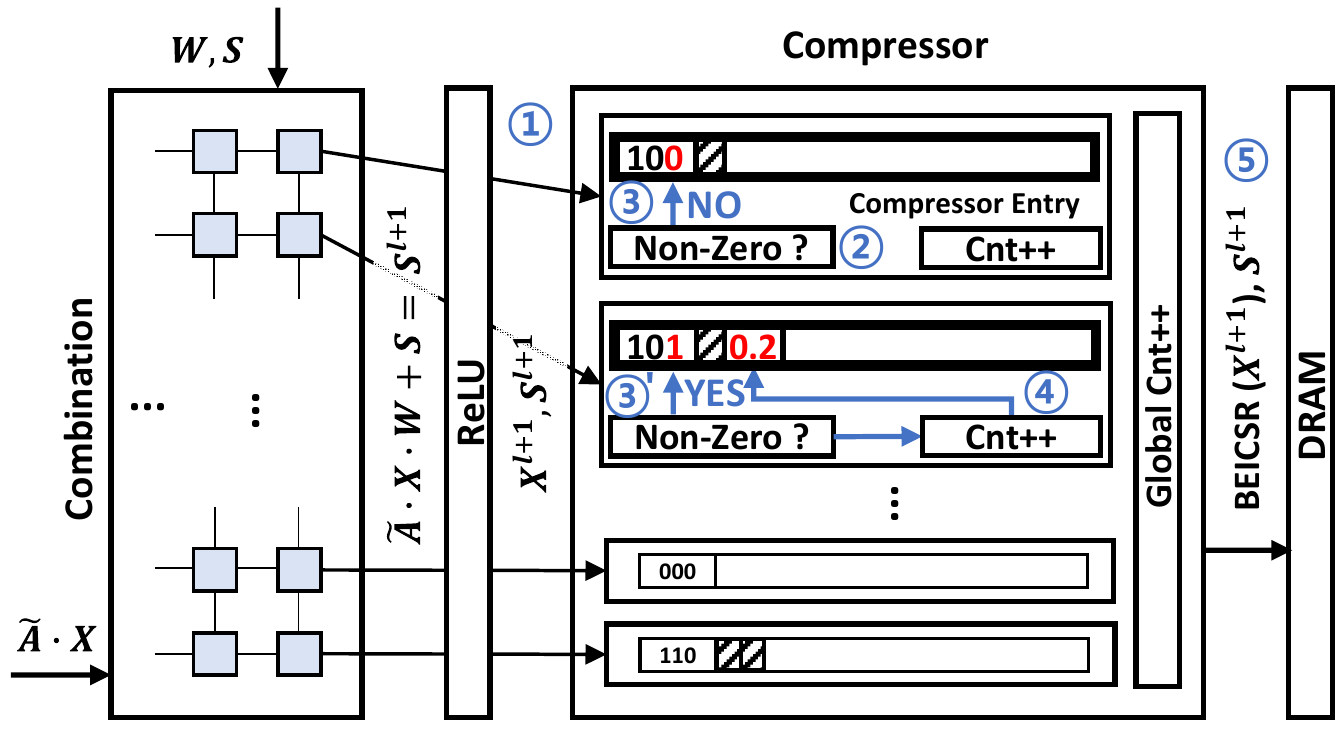}
    \caption{\scheme compressor unit microarchitecture.}%\JL{fix BICSR}
    \label{fig:arch_comp}
\end{figure}

\subsection{Post-Combination Compression}
\label{sec:combcomp}
% \JL{I did a round on this subsec. plz chk}
\cref{fig:arch_comp} illustrates the compressor unit in \scheme and the compression process.
To avoid the extra memory access for the compression, we place the compressor at the output stage of the combination engine using an output-stationary systolic array.
One compressor entry containing \format buffer and compression logic is assigned to each row of the systolic array.
\circled{1} When the combination phase is complete, the data are streamed to the compressor after processing the residual addition and the ReLU activation.
\circled{2} The compression logic checks whether the output value is zero.
\circled{3} If the output value is zero, the bitmap index in the compression entry appends a `0'.
$\circled{3}^\prime$ When the output value is non-zero, the bitmap index accumulates a `1', and \circled{4} the output value is saved to the location pointed by the counter. %counter++
The compressor continuously performs \circled{1}-\circled{4} for each output value from the systolic array.
\circled{5} After the compressor has processed a unit slice amount of data, the data stored in the buffer is flushed to the DRAM, and the compressor is re-initialized.

% \begin{figure}
% \centering
%  \includegraphics[width=0.7\columnwidth]{figures/lac_split.pdf}
%  \caption{\Lac effectively splits a graph into multiple smaller graphs of similar shapes. }
%  \label{fig:lacshape}
% \end{figure}

\subsection{Putting It All Together}

\begin{figure}[t]
    \centering
    \includegraphics[width=\columnwidth]{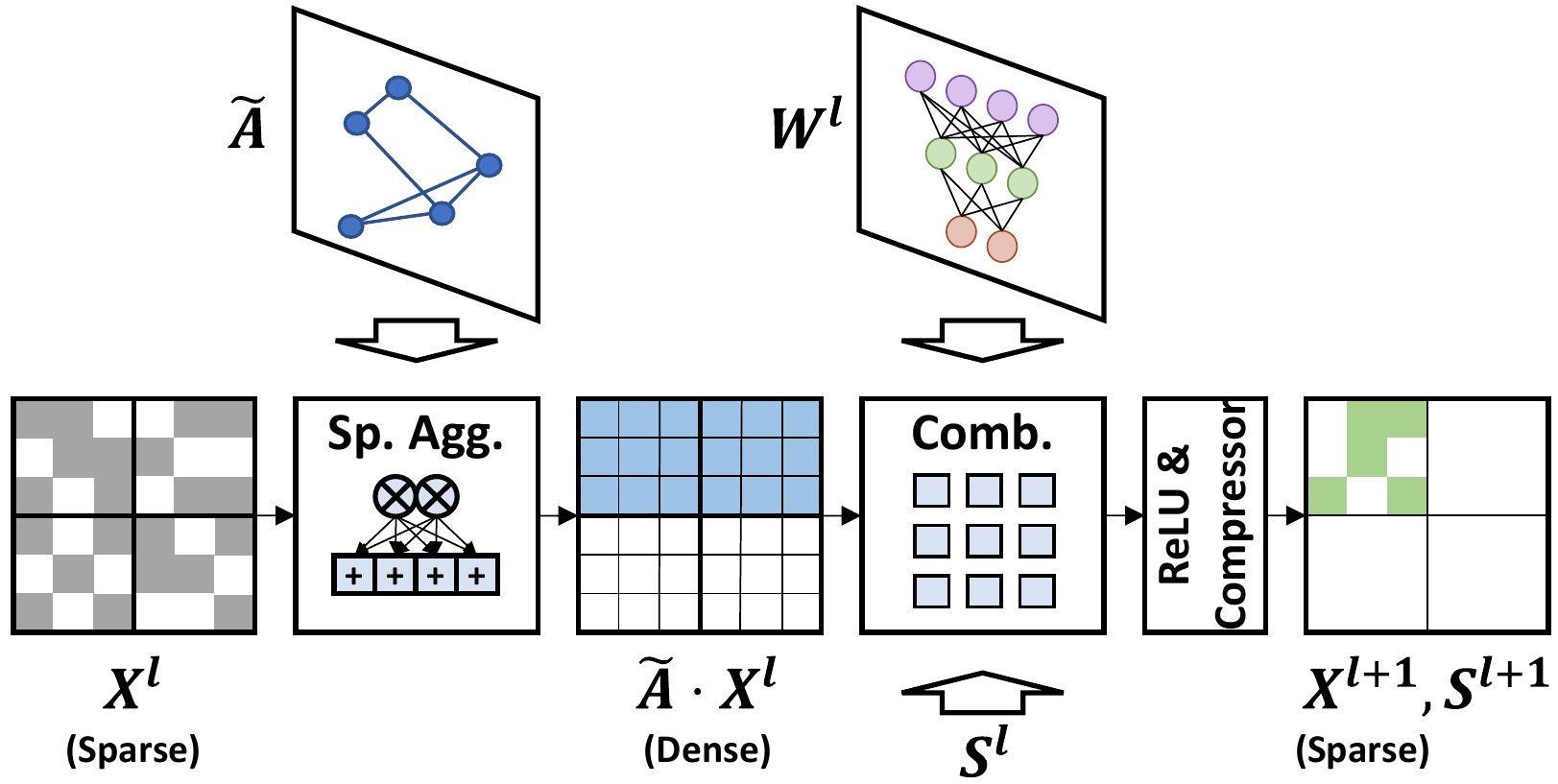}
    \caption{\scheme execution flow for a GCN layer. }
    \label{fig:together}
\end{figure}

\cref{fig:together} presents the overall procedure of \scheme on top of the baseline architecture ( \cref{fig:baseline:arch}).
The sparse aggregator unit takes \A, the graph topology in CSR format, and \Xl in \format format from the output of the previous layer.
Similar to \cite{hygcn, engn, Yoo2021Making}, we use row-product-based dataflow, and apply tiling for both the graph topology \A and feature \Xl as in \cite{engn, Yoo2021Making, gcnax}.
The resulting \AXl will be dense, because each row of \AXl will be a weighted sum of several rows from $\mathit{X^l}$.

After a block of \AXl is calculated from the sparse aggregator unit, it will be sent to the systolic-array-based combination engine. 
The systolic array will multiply \AXl with \Wl, which is essentially a GeMM operation.
To perform the residual addition, the registers at the systolic arrays are initialized with $S^l$ instead of zero.
When a sliced row of \AXlWl is calculated, it is fed to the compressor unit after being activated by ReLU.
The output of ReLU activation is the next layer's input feature $X^{l+1}$.
Before being written to the memory, the post-combination compression unit converts the row of $X^{l+1}$ into the \format format.
In modern GCN architectures, the input and output feature widths are often the same~\cite{gcnii, thousandgcn}. 
In such cases, \Xprime consumes the same capacity with \AX, and compressed \Xprime is written in place of \AX to reduce the memory requirement.

Optionally, when the input feature ($X^1$) is extremely sparse,
the combination of the first layer is performed on the sparse aggregator engine instead to take advantage of the sparse input features. 
Even though this technique only applies to the first layer, this brings a meaningful speedup to some datasets.
\rev{SGCN does not require much additional implementation cost from the baseline GCN accelerator.
%The implementation cost of \scheme is almost negligible.
For the sparse aggregator units, only the additional prefix sum units are required for reading the bitmap index.
For the \scheme compressor units, the global counter and non-zero comparators are added to the basic combination unit to write the bitmap index in addition to the features. %, which do not cause much hardware overhead.
}

% \begin{figure}[t]
%     \centering
%     \includegraphics[width=\columnwidth]{figures/feat_sensi.png}
%     \caption{\scheme feature width sensitivity.}
%     \label{fig:featsensi}
% \end{figure}

\section{Evaluation}

\subsection{Experimental Setup}
\label{sec:setup}

\begin{table}[b]
\scriptsize
\small
\centering
\caption{\rev{Benchmark Dataset Information}}
     \resizebox{\columnwidth}{!}
    {
\begin{tabular}{cccccccc}
\toprule
Dataset & \#Vertices & \#Edges  & \makecell{\#Input\\Features} & Topology & Feature & \makecell{Sparsity \\(w/ 28 layers)} & \makecell{\rev{Accuracy} \\(w/ 28 layers)} \\
\midrule
%    \cmidrule(lr){1-1}
%    \cmidrule(lr){2-5}
Cora (CR) & 0.003 M & 0.01 M & 1,433 &  0.09 MB  &   0.01 GB & 66.1 \% & 76 \% \\
CiteSeer (CS) & 0.003 M & 0.009 M & 3,703 & 0.08 MB & 0.05 GB & 69.7 \% & 66 \% \\
PubMed (PM) & 0.02 M & 0.09 M & 500 & 0.75 MB & 0.04 GB & 70.7 \% & 77 \% \\
NELL (NL) & 0.07 M & 0.25 M & 61,278 & 2.17 MB  & 15.0 GB & 51.0 \% & 64 \% \\
Reddit (RD) & 232 M & 115 M & 602 & 875 MB  &  0.52 GB & 58.4 \% & 95 \% \\
Flickr (FK) & 0.09 M & 0.90 M & 500 & 7.21 MB & 0.17 GB & 46.5 \% & 48 \% \\
Yelp (YP) & 716 M & 14.0 M & 300 & 109 MB & 0.80 GB & 64.0 \% & 54 \% \\
% Amazon-Photo (AP) & 7.7 K & 0.24 M & 745 & 1.85 MB & 21.7 MB & 55.3 \% \\
DBLP (DB) & 0.02 M & 0.11 M & 1,639 & 0.87 MB & 0.11 GB & 59.5 \% & 86 \% \\
\rev{GitHub (GH)} & 0.04 M & 0.58 M & 128 & 4.55 MB & 0.02 GB & 44.6 \% & 86 \% \\ 
% YouTube (YT) &1.13 M&2.99 M&256&0.02 GB  &  1.08 GB \\ 
% Pokec (PK) &1.63 M&30.6 M&256 & 0.12 GB  &  1.56 GB \\
% Wiki (WK) &1.79 M&28.5 M&256 & 0.22 GB & 1.71 GB \\
% Citation (CT) & 2.93M& 30.5M&128 & 0.12 GB & 2.79 GB \\
% Products(PD) & 2.45M& 61.9M & 100 & 0.24 GB & 0.91 GB \\
%WikiTalk (WT) &2.39 M&5.02 M&256& 0.03 GB  &   2.28 GB \\ %wiki-topcats
%LiveJournal (LJ) &4.85 M&69.0 M&256 & 0.28 GB  &   4.63 GB \\
%Orkut (OK) &3.07 M&117 M&256 & 0.45 GB  &  2.93 GB \\
%Reddit (RD) &0.23 M&114 M&256 & 0.43 GB  &  0.22 GB \\
 \bottomrule
\end{tabular}
}
\label{tbl:dataset}
\end{table}

\begin{table}[t!]

% \scriptsize
%\small
% \footnotesize
\centering

\caption{System Configuration}
\begin{tabular}{cccc}
\toprule
%\multicolumn{3}{c}{\textbf {Common}}\\
\multirow{3}{*}{\makecell{Accelerator Engine}} & Frequency & 1GHz \\
& Combination & 32$\times$32 Syst. Array \\
& Aggregation & 16-Way SIMD \\
\cmidrule(lr){2-3}

 \multirow{2}{*}{\#Engines} & Aggregation & 8 \\
& Combination & 8\\
\cmidrule(lr){2-3}

\multirow{3}{*}{Global Cache} & Capacity & 512KB \\
& Ways & 16 \\
& Replacement & LRU \\
\cmidrule(lr){2-3}

\multirow{4}{*}{\makecell{Off-chip Memory}} & Spec. & HBM2 \\
& Peak Bandwidth & 256 GB/s \\
& Channels & 8 \\
& Banks & 4$\times$4 \\

 \bottomrule
\end{tabular}
\label{tbl:system}
\end{table} 

\begin{figure*}[t]
    \centering
    \includegraphics[width=.8\textwidth]{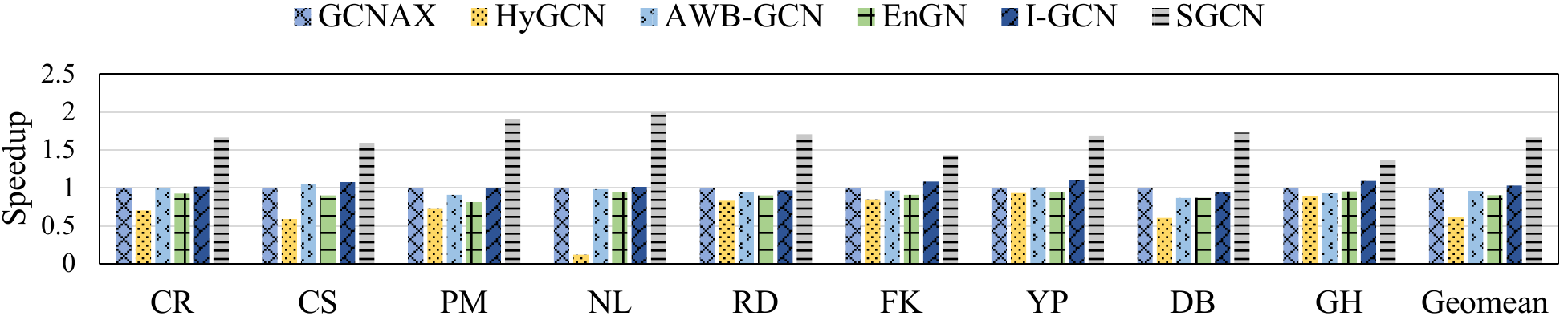}
    \caption{\rev{Performance comparison of GCN accelerators.}}
    \vspace{-3mm}
    \label{fig:sgcnabl}
\end{figure*}

\JL{maybe we wanna say the four datasets (FK-DB) are for low-sparsity examples}
We chose \rev{nine} real-world graph datasets from various sources, listed in \cref{tbl:dataset}.
As sparsity of the actual trained model is essential for \scheme, we trained 28-layer GCNs with residual connection and 256 features per vertex by these datasets based on practices of~\cite{deepgcn, deepergcn}.

Cora (\emph{CR)}, CiteSeer (\emph{CS}), PubMed (\emph{PM}), DBLP (\emph{DB}) are citation network graphs from~\cite{coraciteseerpubmed, dblp}.
Each node (document) is linked with others and has a feature vector consisting of bag-of-words.
% Each node is one of 2,708 documents, associated with 1,433 distinct word vocabulary.
NELL (\emph{NL)}~\cite{nell} is a knowledge graph from the Never-Ending Language Learning project. 
Each node has a one-hot vector of length 61,278.
Reddit (\emph{RD})~\cite{reddit} is a graph often used for GCN kernel evaluation.
Flickr (\emph{FK)}~\cite{snap} is an image relationship graph, and each node has the description and properties of an image as features.
Yelp (\emph{YP}) is a social relationship graph that has customer reviewers as nodes and their relationships as a link.
For YP, we followed ~\cite{graphsaint-iclr20} while preprocessing.
\rev{GitHub (\emph{GH})~\cite{musae} is a code hosting relationship graph.}
% Amazon-Photo (\emph{AP})~\cite{amznphoto,shchur2018pitfalls}.
%DBLP (\emph{DB})~\cite{@} is a citation .
We used original features from these datasets. % for training.

% Citations (\emph{CT}) and Products (\emph{PD}) are from the Open Graph Benchmark~\cite{ogb}.
% The vertices of this graph have corresponding documents (product descriptions for PD and paper abstracts for CT), and the input features are embeddings generated from these documents.
% Pokec (\emph{PK}), YouTube (\emph{YT}), and Wiki (\emph{WK}) are from \cite{snap} and represent large-scale real-world graphs.
% We used Node2vec~\cite{node2vec} to generate 256-dimensional embeddings for each vertex as input features.
%Lastly, Reddit (\emph{RD})~\cite{reddit} is a graph often used for GCN kernel evaluation.
%The graph has per-vertex properties, to which we apply Principal Component Analysis~\cite{pca} to map it to 256 dimensions for input features.

We performed the evaluation using the accelerator architecture explained in \cref{sec:acc}, \cref{sec:spagg} and \cref{sec:combcomp}.
The system configurations are summarized in \cref{tbl:system}.
Following \cite{gcnax}, we used a 512KB global cache unless otherwise stated.
The accelerator runs at 1~GHz. 
The accelerator has 32$\times$32 systolic array-based combination engines and the sparse aggregation engines from \cref{sec:spagg}, \rev{with 32bit fixed points for both features and weights.} 

We designed the accelerator using Verilog HDL and synthesized it using Synopsys Design Compiler with the 45~\SI{}{\nano\meter} OpenPDK library and scaled it to 32~\SI{}{\nano\meter} to match the tech node of the on-chip memory model. 
For the on-chip caches, we used CACTI 6.5~\cite{cacti} to estimate the area and power consumption because of the lack of a memory cell library.
The baseline chip area of GCNAX is 3.95~\SI{}{\square\milli\meter}, 
and the area of the accelerator with \scheme is  4.05~\SI{}{\square\milli\meter}, 
which has 2.5\% overhead on the area, accounting for the increased logic in sparse aggregation and compression engines. %\NK{compression engine?} % and power, respectively.
We validated that it functioned correctly and drew the area and power consumption values.
For comparison, our reproduced AWB-GCN consumed 4.25 \SI{}{\milli\meter\squared} due to the complicated logic.

To measure performance, we designed a cycle-accurate simulator written in C++. 
The model is based on SCALE-Sim~\cite{scalesim} for the systolic array used in the combination engine, 
which we extended with an in-house module for the sparse aggregation engine and compression engine. %\NK{compression engine?}
The execution cycle of the computation modules has been validated with the HDL design at the cycle level.
For modeling the DRAM subsystem, we use DRAMsim3~\cite{dramsim3} with an HBM2 module.

\subsection{Fast and Energy-Efficient GCN Executions}
\label{sec:fast}

\textbf{Performance.} \cref{fig:sgcnabl} demonstrates that \scheme outperforms existing GCN accelerators by a great margin. 
In \cref{fig:sgcnabl}, the performance of \scheme is compared with the previous methodologies. %\NK{to the methodology?} 
HyGCN~\cite{hygcn} is based on row product execution with hybrid engines.
EnGN~\cite{engn} uses vertex tiling and degree-aware vertex caching (DAVC) to cache high-degree vertices explicitly. 
AWB-GCN~\cite{awb} employs column-product-based execution and applies aggressive load-balancing techniques.
In addition, I-GCN~\cite{igcn} adopts dynamic reordering to enhance the locality of the graph topology.
Moreover, GCNAX~\cite{gcnax} is based on perfect tiling and suggests optimized loop ordering based on off-line analysis.
Among those, we choose GCNAX as the baseline and report normalized speedups because it shows successful performance over existing accelerators and uses the most similar tiling approach to \scheme. %\NK{why? most similar approach?} \JS{I added explanations.}

The results show that \scheme achieves 1.66$\times$ speedup over baseline GCNAX in geometric mean, and 2.71$\times$ over HyGCN. %\NK{1.71x? consistency with 2.91x} \MG {I will fix them}
%Comparing with GCNAX helps analyzing the results, 
Because GCNAX employs aggressive tiling for both the topology (\A) and the intermediate features (\X),
the speedup of \scheme is mostly from exploiting the feature sparsity. %\NK{compare GCNAX first. Matching the order with graph makes easier to follow \& picked GCNAX as baseline} \MG {I changed the order}
The speedup over HyGCN is mainly coming from two factors: Reduced amount of DRAM accesses due to the usage of \format format, and cache efficiency from graph/feature tiling. 
Because HyGCN does not perform any tiling/slicing, it suffers from a low cache efficiency for large graphs. %\NK{for large graph?} \MG {I added it }
AWB-GCN adopts zero skipping on the features, which is a kind of sparsity-aware method.
However, AWB-GCN stores the features in a dense format that yields no benefit to the memory traffic. 
Moreover, it uses the sparsity of the features only in the combination phase which only takes a small portion of the total GCN execution. 
%Therefore, its gain becomes marginal. 
By adopting a sparse format to reduce the memory traffic, \scheme outperforms AWB-GCN by 1.73$\times$ in geometric mean.
\scheme also achieves 1.85$\times$ speedup compared to EnGN.
The degree-aware vertex cache used in EnGN is effective over HyGCN, but its limited vertex tiling still makes lower cache efficiency, so \scheme has an extra advantage over EnGN.
% In our benchmark dataset, the degree-aware vertex cache used in EnGN was found effective only on a few datasets \NK{which datasets? seems not effective at all in the graph}, and it was outperformed by \scheme due to its limited vertex-only tiling.
%\MG {EnGN description is needed : degree-aware vertex cache was effective but its limited vertex only tiling still makes lower cache efficiency.}\JL{@MG would you put one here?}\JS{@MG I added brief sentence.}

Over the baseline, a large speedup was observed on PubMed and NELL dataset with 1.91$\times$ and 1.99$\times$ improvement over GCNAX, respectively. 
The PubMed dataset exhibits high intermediate feature sparsity of almost 70\% (see \cref{fig:moti_sparsity_analysis}), which translates to a high speedup in the aggregation phase.
On the other hand, NELL dataset shows relatively lower feature sparsity.
However, its input feature width is exceptionally long (61,278) and is also ultra-sparse (99.9\%) because the input features are one-hot encoded vectors. %\NK{number?} \MG {I added} \NK{input feature of the first layer?} \MG {yes. we call features of hidden layers to intermediate features}
% Thus, the speedup coming from the input layer contributes to the speedup in addition to the intermediate features. 
Thus, this unique input layer brings additional speedup, besides the speedup from the intermediate features.
%\NK{how many layers in NELL?} \MG {We set 28 layers for all of graph, 61278/256 = 240; it is super overhead.}
On the other hand, speedup numbers from Cora and CiteSeer are similar to the geomean, despite its relatively high sparsity. 
The reason is partially from the small dataset.
When the graph topology is small, the relative portion of the combination phase increases, which amortizes the gain coming from the sparse aggregation.
In addition, the average degrees of the two graphs are very low with 3.92 for Cora and 2.76 for CiteSeer where geomean is 10.15 and the maximum is around 500 with Reddit. \NK{compared to XX for XX}This means that the number of random accesses to the features is smaller than the other datasets, which also contributes to the amortization of the gain in sparse aggregation.

\JL{Maybe I can add MG's interpretation on flicker's low sparsity here}
% The reason can be found from low average degree of the two graph topology. 
% In contrary to other datasets with average degree 

% \begin{figure}[t]
%     \centering
%     \includegraphics[width=\columnwidth]{figures/format_ablation.pdf}
%     \caption{Effect of \format and Slicing.\JS{Merge fig15 and this figure into a single figure. todo-Instead replace this figure with layer number sensitivity.}}
%     \label{fig:bicsrablation}
% \end{figure}

\begin{figure}[t]
    \centering
    \includegraphics[width=\columnwidth]{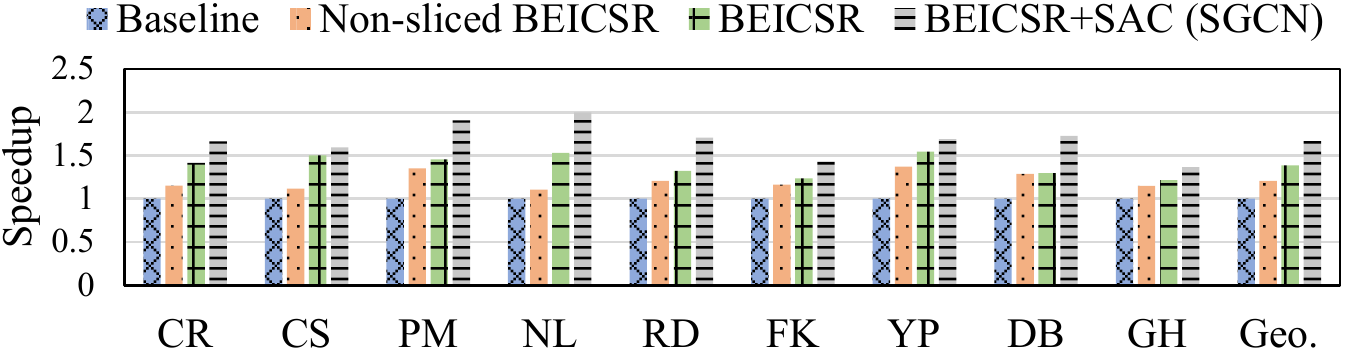}
        \caption{\rev{Ablation study.}}
        \vspace{-3mm}
    \label{fig:ablation}
\end{figure}

\textbf{Ablation Study.}
\cref{fig:ablation} displays how each proposed technique contributes to the performance gain of \scheme.
Using GCNAX as the baseline, we first apply the non-sliced version of \format. 
The non-sliced version of \format is already enough to exploit the intermediate feature sparsity, but settles at suboptimal dataflow due to the lack of feature matrix slicing. %\JL{I have to guess on the results based on fig14. ping me when the ablation is ready.}
As a result, the performance gain is often not large enough.
% , and sometimes negative where the optimal dataflow requires aggressive feature matrix slicing (e.g., AP dataset).
% \MG {I removed the sentence about negative non-sliced BEICSR performance. because it was only on AP.}
The geometric mean performance gain is 20.8\%.

When the feature matrix slicing is supported with the sliced version of \format, sparsity is exploited on the optimal dataflow.
The geometric mean speedup is 38.5\% from the baseline, which adds 17.7\% on top of the non-sliced \format.

Lastly, we add \lac to better capture the varying locality of the sparse features.
This adds the extra 28.0\% speedup in the geometric mean, which results in an overall 1.66$\times$ speedup. %\NK{how much from baseline?} \MG {70.7\% (1.71$\times$). it mentioned on \ref{sec:fast}} \JS{I changed the sentence.}
\Lac adds more gain in graphs with more clustered topology (DB) and high neighbor similarity (PM, RD).

%[multi-engine vs single engine. deprecated]
% Observing the performance from the single-engine accelerators (\cref{fig:sgcnabl} (right)) gives some additional insight on how the performance gain was achieved.
% In our setting, the single-engine alone is not sufficient to consume all the bandwidth from the HBM2 memory (256GB/s).
% Therefore, in this setting, the workloads become compute-bounded.
% An easy observation is that GCNAX does not have much speedup over HyGCN, in contrast to that of the multi-engine setting. 
% The main strategy of GCNAX is to tile the matrices, such that the memory bandwidth requirement can be reduced.
% However, with an abundant amount of bandwidth available for a single-engine to use, the memory traffic reduction provides little benefit.
% With the equal number of arithmetic operations (MACs), almost no speedup is gained for GCNAX over HyGCN (< 1.0\%).
% On the other hand, \scheme requires less number of arithmetic operations to execute. 
% Even though most of the speedup of \scheme comes from the memory access reduction, 
% it still has a meaningful gain of 14.1\% on a single engine over GCNAX, by avoiding the multiplications with zero-values features.

\begin{figure}[t]
    \centering
    \includegraphics[width=\columnwidth]{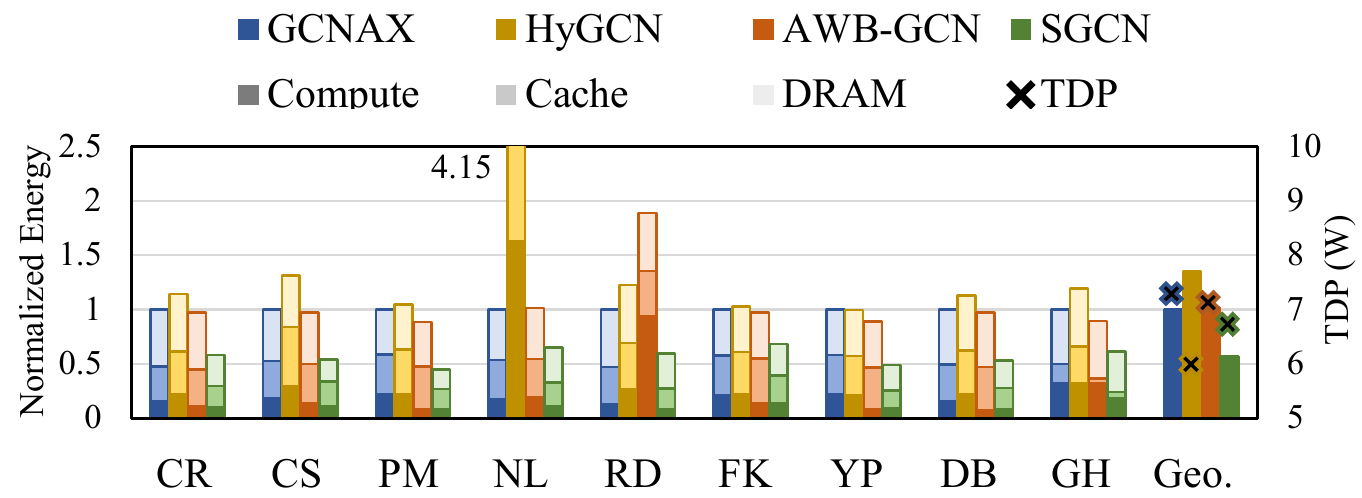}
    \caption{\rev{\scheme energy consumption breakdown.} }%\NK{check cationsss}
    \label{fig:energy}
\end{figure}

\textbf{Energy Consumption.}
\NK{AFAIK, GCN is memory intensive not compute intensive, but the breakdown shows most energy consumption is from the computation.}
\cref{fig:energy} shows the advantage of energy consumption.
\scheme consumes 44.1\% less energy compared to GCNAX, 44.6\%, and 58.1\% compared to AWB-GCN and HyGCN, respectively.
%Compared to the baseline GCNAX, t
The energy savings come from all three parts of the stack.
Much of the energy consumption comes from memory accesses.
Because the aggregation phase requires fewer data due to the sparsity, it affects both the off-chip memory and cache access counts.
In addition, the reduced number of multiplications in the aggregation contributes to the energy reduction of the computation.
\rev{
For the peak power, SGCN consumes 6.74\si{\watt}, which is less than AWB-GCN (7.03\si{\watt}) and GCNAX (7.16\si{\watt}).
However, it shows higher peak power consumption compared to HyGCN (5.94\si{\watt}) which has slow but simple architecture.}

\NK{computation? not DRAM? This paper is claiming reduction of memory access in design goal}
% \scheme saves most of energy from the computation. 
% Because \scheme requires less amount of feature data in aggregation phase.
% it consumes 45.3\% less computation energy, mainly from the reduced amount of multiplications in aggregation phase.
% Moreover, it leads to less amount of DRAM access and read/write access to the feature cache. \MG {writing is too hard.. please check this part.}
% \JS{Should be fixed...}

% Because \scheme loads less amount of feature data, much savings come from the computation and DRAM access.
% Similarly, it leads to less amount of read/write accesses to the feature cache.
% The savings from computation is rather small, because the computation energy is mostly spent on the combination and \scheme consumes a little extra energy for decoding the \format format.
% Nonetheless, it consumes 2.0\% less computation energy, mainly from the reduced amount of multiplications in aggregation phase. \JL{see if above are true when the breakdown comes out} \MG{I will change the description. because of feature compression, computation can save most of energy - 45.0\% compare to computation, 29.3\% compare to total (66.6\% of total saving energy)}

\begin{figure}[t]
    \centering
    \includegraphics[width=1\columnwidth]{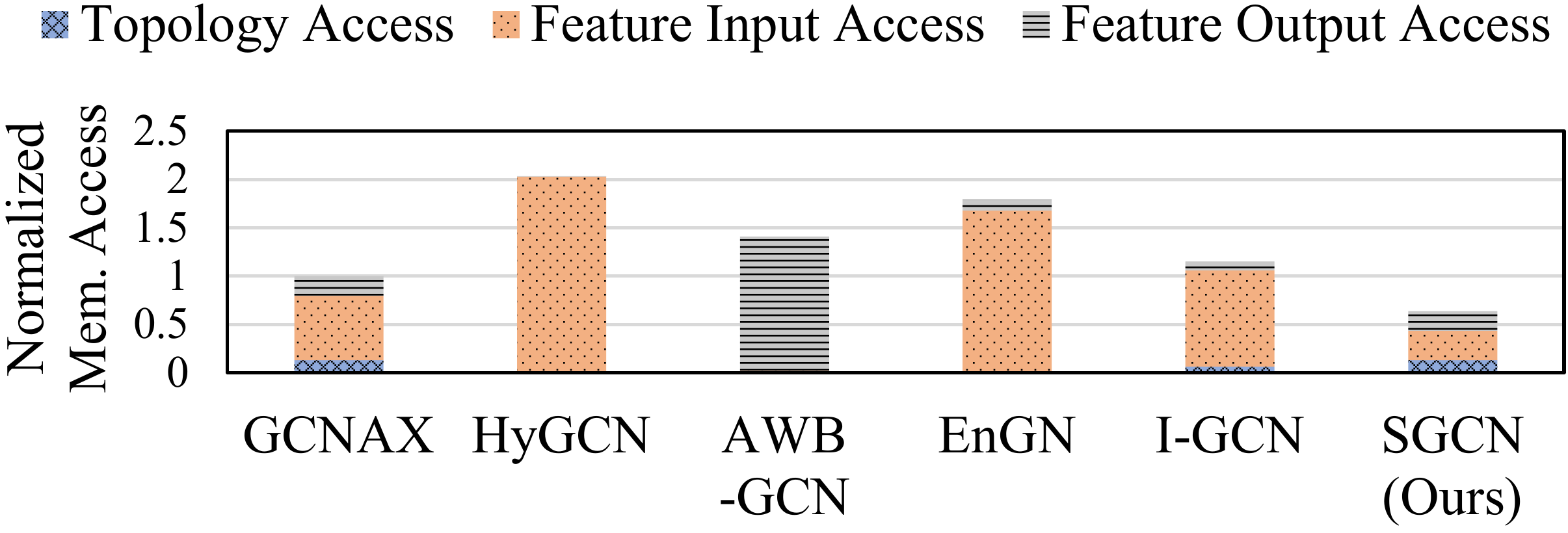}
        \caption{Off-chip memory access breakdown of Reddit (RD).}
    \label{fig:membreak}
\end{figure}

\NK{peformance breakdown?}

\textbf{Memory Access Breakdown.} \cref{fig:membreak} plots the breakdown of the memory accesses during the execution of the RD dataset.
In HyGCN, most of the accesses are to the feature. 
Because it does not use tiling nor utilize the feature sparsity, a lot of duplicate accesses are done to the features which comprise most of the memory access of HyGCN.
EnGN reduced some accesses with its degree-aware vertex cache (DAVC).
On the other hand, AWB-GCN %spends most of its memory access in writing the output.
is based on a column-product dataflow, and it reads each input feature element exactly once.
However, the expense is that it requires reading and writing the partial sums more often, which becomes dominating for the memory accesses.
GCNAX has balanced memory access with aggressive tiling.
I-GCN also shows balanced memory access due to its well-clustered reordering scheme.
%However, GCNAX utilizes the on-chip memories as scratchpad memories, and ends up having to use too small tiles.
%The result is that it consumes much access to the graph topology data.
In contrast, \scheme dramatically reduces the amount of memory access.
The main reduction comes from the sparse feature representation, which reduces the feature access by 54.3\%.
%In addition, using the caches allows for use of larger tiles (at the expense of slightly lower hit rate) and also results in a lower graph topology data access.
%JL{I am not going to endorse the choice of using SRAM as caches}
%\NK{EnGN and I-GCN?}\JS{I added}

\subsection{Sensitivity Studies}
\label{sec:sensi}

\begin{figure}[t]
    \centering
    \includegraphics[width=\columnwidth]{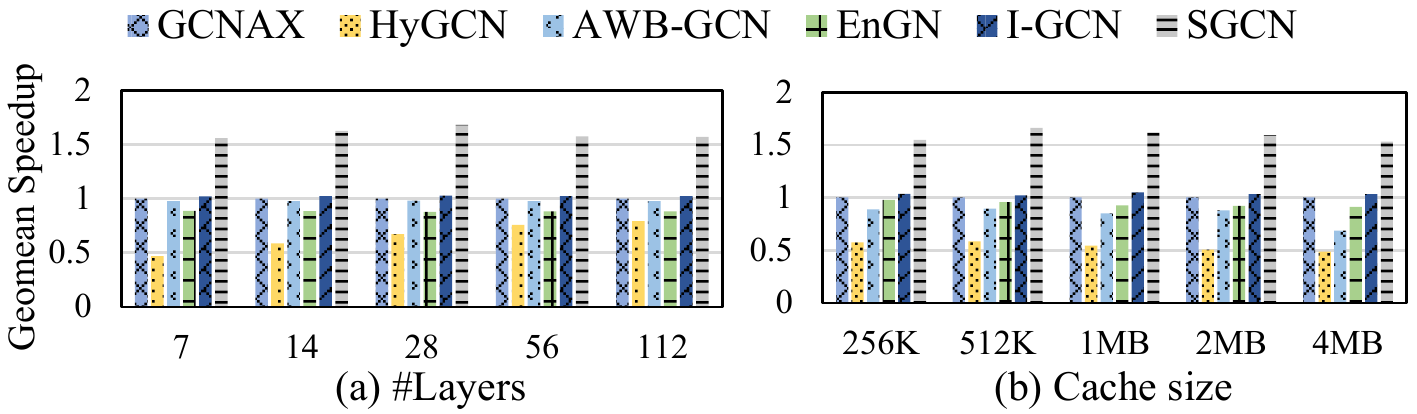}
    \caption{\rev{Sensitivity on the number of layers and cache size.} }
    \label{fig:layers_cache}
\end{figure}

\textbf{Number of GCN Layers.}
In \cref{fig:layers_cache}a we plot the geometric averaged performance sensitivity to the number of layers in GCN using CR, CT, and PM datasets. %\NK{averaged?} \JS{Yes it's geometric mean}
In addition to the default setting of 28 layers, we used 7- to 112-layer GCNs widely used in~\cite{deepgcn, deepergcn}.
%By default, we had been using 28-layer GCN following \cite{deepergcn, deepgcn}.
%\NK{any reason for 7 to 112?}\JS{following the convention in references... i added explanation about it}
In all settings, the sparsity remains mostly constant, and the speedup trend persists.
This shows that the performance gain from \scheme is not fine-tuned on a certain number of layers, and can be broadly used for various specifications.
\NK{Q. How can we determine proper \#layers in GCN? }
\JL{that is an AI question, not architectural issue}

\textbf{Cache Size.} \NK{check}
\cref{fig:layers_cache}b plots the trend of a speedup as the cache size increases. 
In general, the speedup from the sparsity of features is not greatly affected by the cache size unless the data entirely fits into the cache.
With a small cache, the benefit of \lac becomes marginal because it becomes harder to capture any locality.
However, even with larger caches, the speedup remains relatively consistent.

\begin{figure}[t]
    \centering
    \includegraphics[width=\columnwidth]{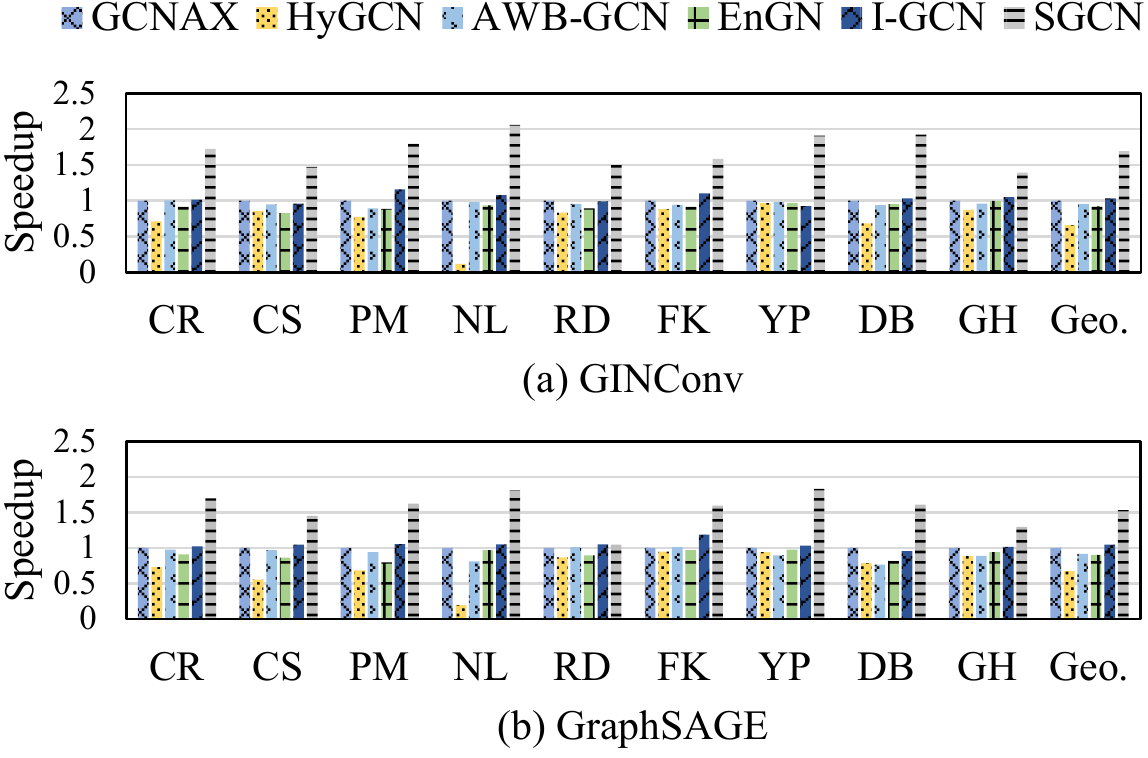}
    \caption{\rev{Performance on (a) GINConv and (b) GraphSAGE.} }
    \label{fig:gnnsensi}
\end{figure}

\textbf{GCN Variants.} \cref{fig:gnnsensi} depicts the performance result comparison on two additional variants of GCN aggregation: GINConv~\cite{ginconv} and GraphSAGE~\cite{graphsage}.
In both variants, \scheme achieves similar but slightly different speedup from those in \cref{fig:sgcnabl}.
Unlike the GCN aggregation~\cite{gcn}, the aggregation phase of GINConv does not require the edge weights.
This decreases the size of \A, and leads to an increase in the portion of the feature matrix during the aggregation.
Because \scheme can efficiently reduce the accesses to the feature matrix, the speedup slightly increases compared to that of the vanilla GCN.
On the contrary, GraphSAGE applies random sampling on the edges to reduce the computational overhead.
It reduces the effective edge count of the graph topology and reduces the portion of aggregation.
Thus, \scheme experiences slightly less, but still significant speedup over the prior arts.
In GINConv, \scheme achieves 1.69$\times$ speedup over GCNAX and 2.57$\times$ against HyGCN.
In GraphSAGE, \scheme achieves 1.53$\times$ and 2.27$\times$ speedup over GCNAX and HyGCN, respectively. %\MG{Also, HyGCN speedup is so large.}
%\JL{needs change. ginconv does not give better speedup}

% \begin{figure}[t]
%     \centering
%     \includegraphics[width=\columnwidth]{figures/scheme_adv.pdf}
%     \caption{Allocated memory according to feature formats and the amount of over-fetching with and without slicing support. \JL{do we need this?} \MG {I think it it not necessary/}}
%     \label{fig:schemeadv}
% \end{figure}

\begin{figure}[t]
    \centering
    \includegraphics[width=\columnwidth]{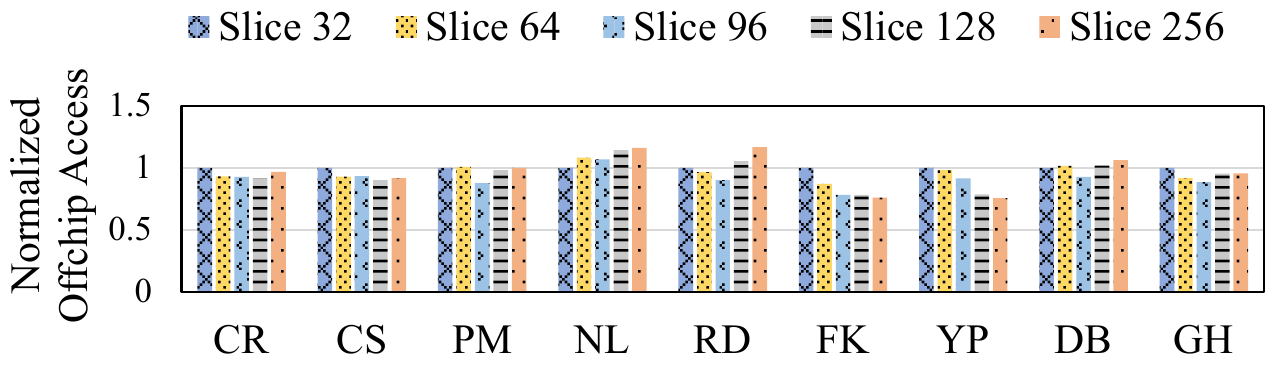}
    \caption{\rev{\scheme slice unit size sensitivity.}}
    \label{fig:slicesensi}
\end{figure}

    % \begin{figure}[t]
    %     \centering
    %     \includegraphics[width=\columnwidth]{figures/single_speedup.pdf}
    %     \caption{Performance comparison on a single engine setting.}
    %     \label{fig:slicesensi}
    % \end{figure}

% \JL{moved?} It is also worth mentioning that naive implementation of sparse aggregation that uses conventional CSRs result in a poor performance, even lower than the prior arts.
% This is because the CSRs do not provide any capacity advantage over the dense format, 
% and the existence of row pointers and separate indices deteriorates drawing the maximum bandwidth out of DRAM, due to the unaligned accesses and frequent random accesses.

\textbf{Choice of Slice Size.}
Another implication of the use of in-place compression with the support for feature matrix slicing is that the speedup can vary depending on the choice of the unit slice size $C$. \JL{S was taken by feature before relu. are we using C anywhere else?}
When the slice width is too large, there is a risk of too many slices occupying the extra cache lines. 
Because this would fetch a lot of invalid data together from the region, it degrades the memory efficiency.
On the other hand, if $C$ is too small, the number of slices increases, which affects the amount of output feature accesses.
Even though the number of extra cache line access will be small, this could increase the execution time.
Nonetheless, \cref{fig:slicesensi} shows that the performance is not very sensitive to the slice size within the range of $C=32$ to $C=256$. % which introduces slightly more over-fetching but exhibits less feature output overhead. %\NK{c=256? s=256?} \JS{S was used in pg.6.}
According to the experiment, the best performance overall is at $C=96$, but a poor choice still provides a great amount of speedup over the baseline. %\NK{largest degradation is about 25\%} \JS{I changed the sentence to compare with GCNAX}%\JL{the best was 96?} \MG{Yes}

\begin{figure}[t]
    \centering
    \includegraphics[width=\columnwidth]{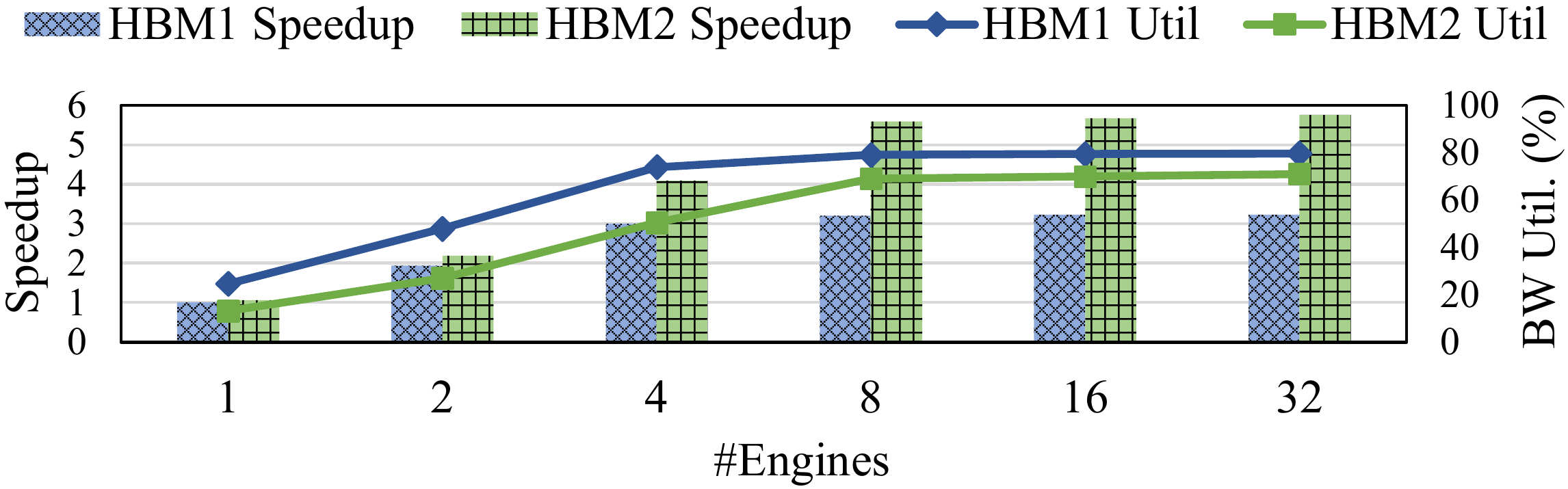}
    \caption{\scheme memory type and scalability. \vspace{-2mm}} %\NK{BW Util.? Util -> Util"." }
    \label{fig:memscale}
\end{figure}

% \begin{figure}[t]
%     \centering 
%     \includegraphics[width=\columnwidth]{figures/lac_effect.pdf}
%     \caption{Effect of LAC on various \#Engines.\JS{Fake!}\JL{maybe we can remove this}}
%     \label{fig:laceffect}
% \end{figure}

\textbf{Scalability.}
\cref{fig:memscale} shows how \scheme scale with the varying number of engines, with different memory modules (HBM1 and HBM2) used.
We distributed the computation to multiple engines.
When enough bandwidth is provided, increasing the number of engines provides an almost linear amount of speedup up to around eight engines, demonstrating good scalability.
The scalability starts saturating at around 16 engines, which is where the system reaches near the maximum bandwidth of the memory module.

\begin{figure}[t]
    \centering
     \includegraphics[width=\columnwidth]{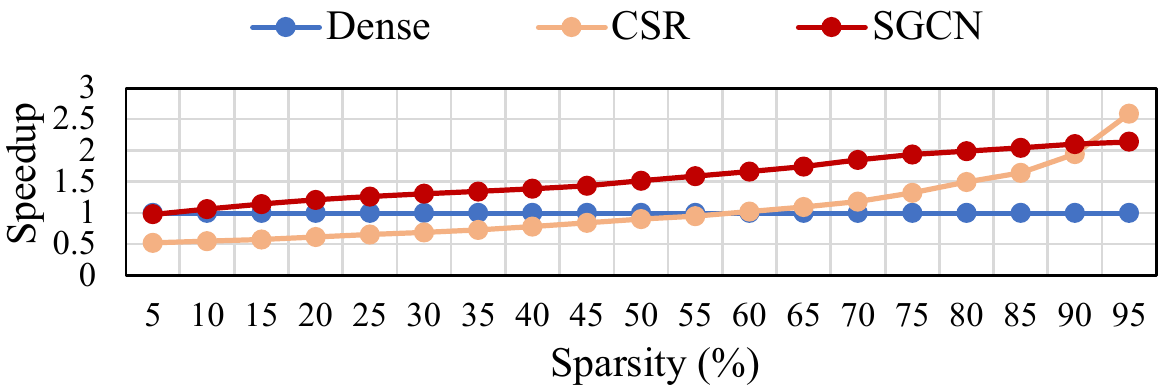}
    \caption{\rev{Performance comparison on various feature sparsity.} }
    \label{fig:sparsitysensi}
\end{figure}

\rev{
\section{Discussion}
\label{sec:discussion}

\subsection{Range of Sparsity}
Based on the observations of \cref{sec:moti}, we designed the \format targeting sparsity around 50\%. 
In fact, all the real-world datasets we examined exhibit the intermediate sparsity between 40\% and 80\%, where we found \scheme to be highly beneficial.
Nonetheless, it should be answered which range of sparsity \format has an advantage over the existing techniques.
Although highly artificial, we plot the geomean speedup of \scheme over various sparsity levels in \cref{fig:sparsitysensi}.
We randomly generated the synthetic input activations of each layer with target sparsity.
As displayed, \scheme shows better performance on almost all sparsity ranges. 
The dense format is better only on sparsity under 5\%, because of the additional bitmap indices needed for \format.
The break-even point for CSR is at over 90\% sparsity, which is the point where the size of the column indices in CSRs becomes smaller than the bitmap indices of \format.
}

\rev{
\subsection{Applicability to Future GCNs}
As demonstrated in \cref{fig:sparsitysensi}, \scheme is beneficial over a wide range of sparsity.
We would like to argue that the future trend would place the sparsity level in a similar range. 
The sparsity of activations is caused by the wide use of ReLU functions.
With normalized values, the after-ReLU distribution will have a near-zero mean, leading to $\sim$50\% sparsity.
Therefore, even if GCNs evolve to have distinct characteristics, the sparsity level is unlikely to change towards the extreme levels.

% of the We dealt with the performance about feature matrix sparsity sensitivity.
% In real world, extremely sparse or extremely dense features are not generated on such a DeepGCN~\cite{deepgcn, deepergcn}.
% TODO\MG{To JS, please fill the reason why it can have extreme sparsity}\\

%\subsection{Alternative Designs on Extreme Case}
One exception is the input layer, which is often human-designed features. 
If the input features of the first layer take a one-hot vector format, sparsity can be extremely high as in NELL~\cite{nell}. 
In this case, using CSR format for input features can be better, and \scheme{} performs combination using the aggregation engine which is designed to handle CSR formats. 
However, the benefit of such a technique is often amortized over tens of layers and hence does not have a significant impact. % on the overall throughput.
}

\section{Related Work}

\subsection{GCN Accelerators}
As GCNs become popular as a critical application domain, the hardware acceleration of GCNs has recently drawn interest from the architecture community. 
HyGCN~\cite{hygcn} claimed the need for GCN accelerators and discussed a few crucial characteristics, such as the hybrid computation pattern and interlayer optimization. 
Moreover, GNNA~\cite{dacgcn} proposed a tiled architecture for GCN acceleration based on such hybrid architectures.
EnGN~\cite{engn} discussed the need for tiling the topology and proposed efficient tile ordering using a dedicated cache for high-degree vertices. 
AWB-GCN~\cite{awb} proposed using column-based execution and applied several load-balancing techniques. %However, these studies do not consider the importance of the repetition structure and miss the opportunity for further optimization. 
GCNAX~\cite{gcnax} is a flexible architecture with an optimal dataflow/buffer configuration found based on perfect tiling. 
However, perfect tiling overprovisions the required amount of buffer and leads to suboptimal performance. 
Moreover, the authors assumed a uniform random distribution of the sparse data, which often does not match the actual distribution. 
SnF~\cite{snf} addresses the issues of perfect tiling and offline analysis from GCNAX by using a dynamic tiling strategy.
By exploiting the repetitive access pattern from feature tiling, it proposes online dynamic and non-perfect topology tiling.
I-GCN~\cite{igcn} suggests a dynamic re-ordering scheme called islandization based on a breadth-first search.
In addition, it reuses the overlapping computations within the aggregation phase to reduce the computational complexity.
A similar approach has been extended in ReGNN~\cite{regnn}, which identifies several redundancies in aggregation and edge updates to further optimize performance.
%
%\begin{sloppypar}
It is also worth noting GCoD~\cite{gcod}, which is a co-design approach for GCNs.
Similar to pruning approaches for DNNs~\cite{pruning}, GCoD prunes, clusters the adjacency matrix, and separately process the dense region and sparse region.
However, such methods require retraining of the GCN, and thus can be used when the user can afford the training.
%\end{sloppypar}

\subsection{Graph Processing Accelerators}

Graph processing forms a vital application domain, and many accelerators have been proposed, based on ASICs~\cite{graphicionado, ozdal, gramer}, FPGAs~\cite{extrav, polygraph, thousand}, or both~\cite{flexminer}.
Because of their high bandwidth requirements, % and random accesses,
graph processing often becomes a primary target for processing-in-memory~\cite{tesseract, graphp, graphq} or processing-in-storage~\cite{graphssd, grafboost}.
%\JL{maybe remove below paragraph}
Tesseract~\cite{tesseract} uses the internal bandwidth of HMCs~\cite{hmc} to boost graph processing, which was later refined with better partitioning~\cite{graphp} or execution alignment~\cite{graphq}.
GraphSSD is an SSD aware of the graph structures, and GraFboost~\cite{grafboost} puts an FPGA accelerator for sort-and-reduce within the storage device for graph processing.  % that is aware
However, they primarily target classic graph processing and are not optimized for GCN execution, especially on the sparsity of the per-vertex features. 

\subsection{Sparse DNNs}
\label{sec:sparsednn}
    
Another stream of related work is on sparse DNN accelerators. 
%
%In modern DNNs, some portion of weight parameters of activations become zero with pruning or using the ReLU activation function.
%Many recent studies on DNN accelerators have been focused on taking advantage of this sparsity.
%
%graph accelerator
%\cite{graphSSD}
%\cite{deepstore} - removed
%\cite{ozdal}
%\cite{grafboost}
%\cite{flexminer} - graph pattern mining
%\cite{polygraph} - fpga flexibility
%\cite{thousand} - cache miss handling graph processing **
%
%
%
% \noindent\textbf{Exploiting the Sparsity in NNs:}
% There are numerous attempts to increase the resource utilization of the NPUs. 
% One stream of work targets sparsity in NNs.
Due to pruning and ReLU activation function, many of the weights or the activations become zeros, providing opportunities for optimization.
Some prior studies design a computation pruning scheme which is friendly to existing hardware ~\cite{Hua2019Boosting, Niu2020PatDNN, Silfa2019Neuron}.
~\cite{Zhu2019Sparse, wang2021dual} are extensions to TensorCores that support sparse operations.
There are also some designs that eliminate zero-value weights~\cite{han2016eie, Zhang2016CambriconX, Zhou2018CambriconS, Albericio2016Cnvlutin, Ding2017CirCNN} or consider zero-value activations~\cite{Parashar2017SCNN, Gondimalla2019SparTen, Judd2017Cnvlutin2, samsung}. 
Some target sparse tensor algebra, targeted to SpMV~\cite{Sadi2019Efficient, Asgari2020ALRESCHA} and SpMM~\cite{sparch, hegde2019extensor, Qin2020SIGMA}.
RingCNN~\cite{ringcnn} uses ring algebra for algebraic sparsity.
In addition, some approaches propose software-hardware co-design of tensor compression~\cite{kanellopoulos2019smash, tensaurus, Pentecost2019MaxNVM, spzip}.
However, these methods primarily target the conventional CNNs, or embedding layers whose characteristics greatly differ from that of GCNs.
To the extent of our knowledge, this is the first work on GCN accelerators that proposes techniques for exploiting sparsity in the features.

%\cite{Qin2020SIGMA} - sparse dnn
%\cite{Kanellopoulos2019SMASH, tensaurus} - compression format w/ acc
%\cite{Zhu2019Sparse, wang2021dual} -sparse tensorcore gpu
%\cite{Pentecost2019MaxNVM} encoding+nvm storage
%\cite{samsung}
%\cite{spzip} = arch compressor for csr
%\cite{ringcnn} - ring algebra?????
%\cite{}
%\cite{}
%zcomp

\subsection{Memory Bandwidth Optimization for DNN Execution}
Memories are known to be important factors of DNN accelerators~\cite{Rhu2016vDNN, Hyun2020NeuMMU, Siu2018Memory}, and optimizations focused on memory bandwidth are often crucial to performance. 
One popular method is to optimize the on-chip memory usage.
Many accelerators adopt different dataflows~\cite{shidiannao, eyeriss, tpu},

and some researchers have claimed that data blocking is more important~\cite{overrated}. 
In a similar sense, some work has focused on the management of on-chip memories~\cite{overcoming, smartshuttle, splitcnn}.
Another direction is to take advantage of the multi-layered nature of DNNs. 
By keeping inter-layer data in on-chip memory, the bandwidth for off-chip traffic can be reduced~\cite{fusedlayer, ecnn, mbs, bnff}.
These methods have been successfully applied to inference~\cite{fusedlayer, ecnn}, training~\cite{mbs}, and batch normalization~\cite{bnff}.
Among many differences, we highlight that these techniques focus on dealing with on-chip memory management, especially as a manually managed scratchpad. 
%These 
methods are valid because of the highly regular memory access pattern in conventional DNNs and fundamentally different from this work.
For GCN execution, irregular accesses from the graphs make it challenging to adopt those common techniques from existing DNN accelerators.

\section{Conclusion}
We proposed \scheme, a GCN accelerator that exploits the sparsity of intermediate features. % for utilizing the memory traffic. %\NK{??}\JS{fixed}
With the advance in GCNs towards deep, residual layers, there appears a huge potential from the intermediate feature sparsity that was unavailable from traditional shallow GCNs. %, the opportunity for the feature sparsity has been overlooked by the extreme sparsity in the graph topology since they are focused on traditional GCNs with a few layers. %\NK{since they are focused on the GCN with a few layers?}

In such circumstances, we present for the first time, exploiting sparsity in the feature data to optimize the memory traffic from the GCN execution.
We first identify that a na\"{i}ve attempt to utilize the feature sparsity could rather result in speed degradation.\JL{not overlooked, but a new opportunity}
From the observation, we present \format, a sparse feature representation format designed for the sparse aggregation phase of the GCN execution.
In addition, we propose \lac, a method for better dealing with the locality in the existence of varying sparsity, with the change in the access patterns.
The key ideas of the format are using embedded bitmap indices, in-place compression, and supporting feature slicing technique which is recently used in GCN accelerators. %\NK{frequently found?} \JS{i think we should remove the word frequently}
By conducting a thorough evaluation, 
We demonstrate that \scheme achieves superior performance and energy efficiency compared to the previous state-of-the-art GCN accelerators.

\section*{Acknowledgements}
This work was partly supported by
% NRF
%  - 신진연구: 2022R1C1C1011307 (JH), 2022R1C1C1008131 (YS)
% IITP
%  - PIM SW: 2021-0-00853
%  - AI대학원: 2020-0-01361
the National Research Foundation of Korea (NRF) grants (2022R1C1C1011307, 2022R1C1C1008131) and
Institute of Information \& communications Technology Planning \& Evaluation (IITP) grants (2021-0-00853, 2020-0-01361) funded by the Korea government (MSIT).
% IDEC
The EDA tool was supported by the IC Design Education Center (IDEC), Korea.
% BK FOUR
Mingi Yoo, Jaeyong Song, Jounghoo Lee, and Youngsok Kim were partly supported by the BK21 FOUR (Fostering Outstanding Universities for Research) funded by the Ministry of Education (MOE, Korea) and National Research Foundation of Korea (NRF).

%%%%%%% -- PAPER CONTENT ENDS -- %%%%%%%%

%%%%%%%%% -- BIB STYLE AND FILE -- %%%%%%%%
% \bibliographystyle{plain}
% \bibliographystyle{plain}
% \balance % don't know why -- but [76] was off the position
\bibliographystyle{IEEEtranS}
\bibliography{refs}

% Generated by IEEEtranS.bst, version: 1.13 (2008/09/30)
\begin{thebibliography}{100}
\providecommand{\url}[1]{#1}
\csname url@samestyle\endcsname
\providecommand{\newblock}{\relax}
\providecommand{\bibinfo}[2]{#2}
\providecommand{\BIBentrySTDinterwordspacing}{\spaceskip=0pt\relax}
\providecommand{\BIBentryALTinterwordstretchfactor}{4}
\providecommand{\BIBentryALTinterwordspacing}{\spaceskip=\fontdimen2\font plus
\BIBentryALTinterwordstretchfactor\fontdimen3\font minus
  \fontdimen4\font\relax}
\providecommand{\BIBforeignlanguage}[2]{{%
\expandafter\ifx\csname l@#1\endcsname\relax
\typeout{** WARNING: IEEEtranS.bst: No hyphenation pattern has been}%
\typeout{** loaded for the language `#1'. Using the pattern for}%
\typeout{** the default language instead.}%
\else
\language=\csname l@#1\endcsname
\fi
#2}}
\providecommand{\BIBdecl}{\relax}
\BIBdecl

\bibitem{tesseract}
J.~Ahn, S.~Hong, S.~Yoo, O.~Mutlu, and K.~Choi, ``A scalable
  processing-in-memory accelerator for parallel graph processing,'' in
  \emph{ISCA}, 2015.

\bibitem{Albericio2016Cnvlutin}
J.~Albericio, P.~Judd, T.~Hetherington, T.~Aamodt, N.~E. Jerger, and
  A.~Moshovos, ``{Cnvlutin}: Ineffectual-neuron-free deep neural network
  computing,'' in \emph{ISCA}, 2016.

\bibitem{fusedlayer}
M.~Alwani, H.~Chen, M.~Ferdman, and P.~Milder, ``Fused-layer {CNN}
  accelerators,'' in \emph{MICRO}, 2016.

\bibitem{Asgari2020ALRESCHA}
B.~Asgari, R.~Hadidi, T.~Krishna, H.~Kim, and S.~Yalamanchili, ``{ALRESCHA}: A
  lightweight reconfigurable sparse-computation accelerator,'' in \emph{HPCA},
  2020.

\bibitem{thousand}
M.~Asiatici and P.~Ienne, ``Large-scale graph processing on {FPGAs} with caches
  for thousands of simultaneous misses,'' in \emph{ISCA}, 2021.

\bibitem{dacgcn}
A.~Auten, M.~Tomei, and R.~Kumar, ``Hardware acceleration of graph neural
  networks,'' in \emph{DAC}, 2020.

\bibitem{bv}
P.~Boldi and S.~Vigna, ``{The webgraph framework I: Compression techniques},''
  in \emph{WWW}, 2004.

\bibitem{nell}
A.~Carlson, J.~Betteridge, B.~Kisiel, B.~Settles, E.~R. Hruschka, and T.~M.
  Mitchell, ``Toward an architecture for never-ending language learning,'' in
  \emph{AAAI}, 2010.

\bibitem{regnn}
C.~Chen, K.~Li, Y.~Li, and X.~Zou, ``{ReGNN: A Redundancy-Eliminated Graph
  Neural Networks Accelerator},'' in \emph{HPCA}, 2022.

\bibitem{gcnii}
M.~Chen, Z.~Wei, Z.~Huang, B.~Ding, and Y.~Li, ``Simple and deep graph
  convolutional networks,'' in \emph{ICML}, 2020.

\bibitem{flexminer}
X.~Chen, T.~Huang, S.~Xu, T.~Bourgeat, C.~Chung, and Arvind, ``{FlexMiner}: A
  pattern-aware accelerator for graph pattern mining,'' in \emph{ISCA}, 2021.

\bibitem{eyeriss}
Y.-H. Chen, T.~Krishna, J.~S. Emer, and V.~Sze, ``Eyeriss: An energy-efficient
  reconfigurable accelerator for deep convolutional neural networks,''
  \emph{JSSC}, 2016.

\bibitem{polygraph}
V.~Dadu, S.~Liu, and T.~Nowatzki, ``{PolyGraph}: Exposing the value of
  flexibility for graph processing accelerators,'' in \emph{ISCA}, 2021.

\bibitem{Ding2017CirCNN}
C.~Ding, S.~Liao, Y.~Wang, Z.~Li, N.~Liu, Y.~Zhuo, C.~Wang, X.~Qian, Y.~Bai,
  G.~Yuan, X.~Ma, Y.~Zhang, J.~Tang, Q.~Qiu, X.~Lin, and B.~Yuan, ``{CirCNN}:
  Accelerating and compressing deep neural networks using block-circulant
  weight matrices,'' in \emph{MICRO}, 2017.

\bibitem{shidiannao}
Z.~Du, R.~Fasthuber, T.~Chen, P.~Ienne, L.~Li, T.~Luo, X.~Feng, Y.~Chen, and
  O.~Temam, ``{ShiDianNao}: Shifting vision processing closer to the sensor,''
  in \emph{ISCA}, 2015.

\bibitem{molecule}
A.~Fout, J.~Byrd, B.~Shariat, and A.~Ben-Hur, ``Protein interface prediction
  using graph convolutional networks,'' in \emph{NeurIPS}, 2017.

\bibitem{awb}
T.~Geng, A.~Li, R.~Shi, C.~Wu, T.~Wang, Y.~Li, P.~Haghi, A.~Tumeo, S.~Che,
  S.~Reinhardt, and M.~C. Herbordt, ``{AWB-GCN}: A graph convolutional network
  accelerator with runtime workload rebalancing,'' in \emph{MICRO}, 2020.

\bibitem{igcn}
T.~Geng, C.~Wu, Y.~Zhang, C.~Tan, C.~Xie, H.~You, M.~C. Herbordt, Y.~Lin, and
  A.~Li, ``{I-GCN: A Graph Convolutional Network Accelerator with Runtime
  Locality Enhancement through Islandization},'' in \emph{MICRO}, 2021.

\bibitem{girvan2002community}
M.~Girvan and M.~E. Newman, ``Community structure in social and biological
  networks,'' \emph{Proceedings of the national academy of sciences}, 2002.

\bibitem{Gondimalla2019SparTen}
A.~Gondimalla, N.~Chesnut, M.~Thottethodi, and T.~N. Vijaykumar, ``{SparTen}: A
  sparse tensor accelerator for convolutional neural networks,'' in
  \emph{MICRO}, 2019.

\bibitem{graphicionado}
T.~J. Ham, L.~Wu, N.~Sundaram, N.~Satish, and M.~Martonosi, ``{Graphicionado: A
  high-performance and energy-efficient accelerator for graph analytics},'' in
  \emph{MICRO}, 2016.

\bibitem{graphsage}
W.~L. Hamilton, R.~Ying, and J.~Leskovec, ``Inductive representation learning
  on large graphs,'' in \emph{NeurIPS}, 2017.

\bibitem{han2016eie}
S.~Han, X.~Liu, H.~Mao, J.~Pu, A.~Pedram, M.~A. Horowitz, and W.~J. Dally,
  ``{EIE}: Efficient inference engine on compressed deep neural network,'' in
  \emph{ISCA}, 2016.

\bibitem{hegde2019extensor}
K.~Hegde, H.~Asghari-Moghaddam, M.~Pellauer, N.~Crago, A.~Jaleel, E.~Solomonik,
  J.~Emer, and C.~W. Fletcher, ``{ExTensor}: An accelerator for sparse tensor
  algebra,'' in \emph{MICRO}, 2019.

\bibitem{disentangle}
I.~Higgins, D.~Amos, D.~Pfau, S.~Racaniere, L.~Matthey, D.~Rezende, and
  A.~Lerchner, ``Towards a definition of disentangled representations,''
  \emph{arXiv preprint}, 2018.

\bibitem{Hua2019Boosting}
W.~Hua, Y.~Zhou, C.~D. Sa, Z.~Zhang, and G.~E. Suh, ``Boosting the performance
  of {CNN} accelerators with dynamic fine-grained channel gating,'' in
  \emph{MICRO}, 2019.

\bibitem{ringcnn}
C.-T. Huang, ``{RingCNN}: Exploiting algebraically-sparse ring tensors for
  energy-efficient {CNN}-based computational imaging,'' in \emph{ISCA}, 2021.

\bibitem{ecnn}
C.-T. Huang, Y.-C. Ding, H.-C. Wang, C.-W. Weng, K.-P. Lin, L.-W. Wang, and
  L.-D. Chen, ``{eCNN}: A block-based and highly-parallel {CNN} accelerator for
  edge inference,'' in \emph{MICRO}, 2019.

\bibitem{Hyun2020NeuMMU}
B.~Hyun, Y.~Kwon, Y.~Choi, J.~Kim, and M.~Rhu, ``{NeuMMU}: Architectural
  support for efficient address translations in {NPUs},'' in \emph{ASPLOS},
  2020.

\bibitem{samsung}
J.-W. Jang, S.~Lee, D.~Kim, H.~Park, A.~S. Ardestani, Y.~Choi, C.~Kim, Y.~Kim,
  H.~Yu, H.~Abdel-Aziz, J.-S. Park, H.~Lee, D.~Lee, M.~W. Kim, H.~Jung, H.~Nam,
  D.~Lim, S.~Lee, J.-H. Song, S.~Kwon, J.~Hassoun, S.~Lim, and C.~Choi,
  ``Sparsity-aware and re-configurable {NPU} architecture for {Samsung}
  flagship mobile {SoC},'' in \emph{ISCA}, 2021.

\bibitem{splitcnn}
T.~Jin and S.~Hong, ``{Split-CNN}: Splitting window-based operations in
  convolutional neural networks for memory system optimization,'' in
  \emph{ASPLOS}, 2019.

\bibitem{tpu}
N.~P. Jouppi, C.~Young, N.~Patil, D.~Patterson, G.~Agrawal, R.~Bajwa, S.~Bates,
  S.~Bhatia, N.~Boden, A.~Borchers, R.~Boyle, P.-l. Cantin, C.~Chao, C.~Clark,
  J.~Coriell, M.~Daley, M.~Dau, J.~Dean, B.~Gelb, T.~V. Ghaemmaghami,
  R.~Gottipati, W.~Gulland, R.~Hagmann, C.~R. Ho, D.~Hogberg, J.~Hu, R.~Hundt,
  D.~Hurt, J.~Ibarz, A.~Jaffey, A.~Jaworski, A.~Kaplan, H.~Khaitan,
  D.~Killebrew, A.~Koch, N.~Kumar, S.~Lacy, J.~Laudon, J.~Law, D.~Le, C.~Leary,
  Z.~Liu, K.~Lucke, A.~Lundin, G.~MacKean, A.~Maggiore, M.~Mahony, K.~Miller,
  R.~Nagarajan, R.~Narayanaswami, R.~Ni, K.~Nix, T.~Norrie, M.~Omernick,
  N.~Penukonda, A.~Phelps, J.~Ross, M.~Ross, A.~Salek, E.~Samadiani, C.~Severn,
  G.~Sizikov, M.~Snelham, J.~Souter, D.~Steinberg, A.~Swing, M.~Tan,
  G.~Thorson, B.~Tian, H.~Toma, E.~Tuttle, V.~Vasudevan, R.~Walter, W.~Wang,
  E.~Wilcox, and D.~H. Yoon, ``In-datacenter performance analysis of a tensor
  processing unit,'' in \emph{ISCA}, 2017.

\bibitem{Judd2017Cnvlutin2}
P.~Judd, A.~Delmas, S.~Sharify, and A.~Moshovos,
  ``{Cnvlutin\textsuperscript{2}}: Ineffectual-activation-and-weight-free deep
  neural network computing,'' \emph{arXiv preprint}, 2017.

\bibitem{grafboost}
S.-W. Jun, A.~Wright, S.~Zhang, S.~Xu, and Arvind, ``{GraFBoost: Using
  accelerated flash storage for external graph analytics},'' in \emph{ISCA},
  2018.

\bibitem{bnff}
D.~Jung, W.~Jung, B.~Kim, S.~Lee, W.~Rhee, and J.~H. Ahn, ``Restructuring batch
  normalization to accelerate {CNN} training,'' in \emph{SysML}, 2019.

\bibitem{kanellopoulos2019smash}
K.~Kanellopoulos, N.~Vijaykumar, C.~Giannoula, R.~Azizi, S.~Koppula, N.~M.
  Ghiasi, T.~Shahroodi, J.~G. Luna, and O.~Mutlu, ``Smash: Co-designing
  software compression and hardware-accelerated indexing for efficient sparse
  matrix operations,'' in \emph{MICRO}, 2019.

\bibitem{gcn}
T.~N. Kipf and M.~Welling, ``Semi-supervised classification with graph
  convolutional networks,'' \emph{arXiv preprint}, 2016.

\bibitem{maeri}
H.~Kwon, A.~Samajdar, and T.~Krishna, ``Maeri: Enabling flexible dataflow
  mapping over dnn accelerators via reconfigurable interconnects,'' in
  \emph{ASPLOS}, 2018.

\bibitem{pruning}
Y.~LeCun, J.~Denker, and S.~Solla, ``Optimal brain damage,'' \emph{NIPS}, 1989.

\bibitem{extrav}
J.~Lee, H.~Kim, S.~Yoo, K.~Choi, H.~P. Hofstee, G.-J. Nam, M.~R. Nutter, and
  D.~Jamsek, ``{ExtraV: Boosting graph processing near storage with a coherent
  accelerator},'' \emph{pVLDB}, 2017.

\bibitem{leskovec}
J.~Leskovec, J.~Kleinberg, and C.~Faloutsos, ``{Graphs over time: Densification
  laws, shrinking diameters and possible explanations},'' in \emph{KDD}, 2005.

\bibitem{snap}
J.~Leskovec and A.~Krevl, ``{SNAP Datasets}: {Stanford} large network dataset
  collection,'' http://snap.stanford.edu/data, 2014.

\bibitem{thousandgcn}
G.~Li, M.~M{\"u}ller, B.~Ghanem, and V.~Koltun, ``Training graph neural
  networks with 1000 layers,'' in \emph{ICML}, 2021.

\bibitem{deepgcn}
G.~Li, M.~Muller, A.~Thabet, and B.~Ghanem, ``{DeepGCNs}: Can {GCNs} go as deep
  as {CNNs}?'' in \emph{ICCV}, 2019.

\bibitem{deepergcn}
G.~Li, C.~Xiong, A.~Thabet, and B.~Ghanem, ``Deepergcn: All you need to train
  deeper gcns,'' \emph{arXiv preprint}, 2020.

\bibitem{bidirectional}
H.~Li, M.~Yan, X.~Yang, L.~Deng, W.~Li, X.~Ye, D.~Fan, and Y.~Xie, ``Hardware
  acceleration for gcns via bidirectional fusion,'' \emph{IEEE CAL}, 2021.

\bibitem{gcnax}
J.~Li, A.~Louri, A.~Karanth, and R.~Bunescu, ``{GCNAX}: A flexible and
  energy-efficient accelerator for graph convolutional neural networks,'' in
  \emph{HPCA}, 2021.

\bibitem{smartshuttle}
J.~Li, G.~Yan, W.~Lu, S.~Jiang, S.~Gong, J.~Wu, and X.~Li, ``{SmartShuttle}:
  Optimizing off-chip memory accesses for deep learning accelerators,'' in
  \emph{DATE}, 2018.

\bibitem{li2018deeperinsights}
Q.~Li, Z.~Han, and X.-M. Wu, ``Deeper insights into graph convolutional
  networks for semi-supervised learning,'' in \emph{AAAI}, 2018.

\bibitem{dramsim3}
S.~Li, Z.~Yang, D.~Reddy, A.~Srivastava, and B.~Jacob, ``{DRAMsim3: A
  cycle-accurate, thermal-capable DRAM simulator},'' \emph{IEEE CAL}, 2020.

\bibitem{engn}
S.~Liang, Y.~Wang, C.~Liu, L.~He, L.~Huawei, D.~Xu, and X.~Li, ``{EnGN}: A
  high-throughput and energy-efficient accelerator for large graph neural
  networks,'' \emph{TC}, 2020.

\bibitem{mbs}
S.~Lym, A.~Behroozi, W.~Wen, G.~Li, Y.~Kwon, and M.~Erez, ``Mini-batch
  serialization: {CNN} training with inter-layer data reuse,'' in \emph{SysML},
  2019.

\bibitem{graphssd}
K.~K. Matam, G.~Koo, H.~Zha, H.-W. Tseng, and M.~Annavaram, ``{GraphSSD: Graph
  semantics aware SSD},'' in \emph{ISCA}, 2019.

\bibitem{Niu2020PatDNN}
W.~Niu, X.~Ma, S.~Lin, S.~Wang, X.~Qian, X.~Lin, Y.~Wang, and B.~Ren,
  ``{PatDNN}: Achieving real-time {DNN} execution on mobile devices with
  pattern-based weight pruning,'' in \emph{ASPLOS}, 2020.

\bibitem{ozdal}
M.~M. Ozdal, S.~Yesil, T.~Kim, A.~Ayupov, J.~Greth, S.~Burns, and O.~Ozturk,
  ``Energy efficient architecture for graph analytics accelerators,'' in
  \emph{ISCA}, 2016.

\bibitem{Parashar2017SCNN}
A.~Parashar, M.~Rhu, A.~Mukkara, A.~Puglielli, R.~Venkatesan, B.~Khailany,
  J.~Emer, S.~W. Keckler, and W.~J. Dally, ``{SCNN}: An accelerator for
  compressed-sparse convolutional neural networks,'' in \emph{ISCA}, 2017.

\bibitem{scnn}
A.~Parashar, M.~Rhu, A.~Mukkara, A.~Puglielli, R.~Venkatesan, B.~Khailany,
  J.~Emer, S.~W. Keckler, and W.~J. Dally, ``Scnn: An accelerator for
  compressed-sparse convolutional neural networks,'' \emph{ISCA}, 2017.

\bibitem{hmc}
J.~T. Pawlowski, ``Hybrid memory cube ({HMC}),'' in \emph{Hot Chips}, 2011.

\bibitem{Pentecost2019MaxNVM}
L.~Pentecost, M.~Donato, B.~Reagen, U.~Gupta, S.~Ma, G.-Y. Wei, and D.~Brooks,
  ``{MaxNVM}: Maximizing {DNN} storage density and inference efficiency with
  sparse encoding and error mitigation,'' in \emph{MICRO}, 2019.

\bibitem{Qin2020SIGMA}
E.~Qin, A.~Samajdar, H.~Kwon, V.~Nadella, S.~Srinivasan, D.~Das, B.~Kaul, and
  T.~Krishna, ``{SIGMA}: A sparse and irregular {GEMM} accelerator with
  flexible interconnects for {DNN} training,'' in \emph{HPCA}, 2020.

\bibitem{qureshi_adaptive_insertion}
M.~K. Qureshi, A.~Jaleel, Y.~N. Patt, S.~C. Steely, and J.~Emer, ``Adaptive
  insertion policies for high performance caching,'' in \emph{ISCA}, 2007.

\bibitem{Rhu2016vDNN}
M.~Rhu, N.~Gimelshein, J.~Clemons, A.~Zulfiqar, and S.~W. Keckler, ``{vDNN}:
  Virtualized deep neural networks for scalable, memory-efficient neural
  network design,'' in \emph{MICRO}, 2016.

\bibitem{compressing_dma}
M.~Rhu, M.~O'Connor, N.~Chatterjee, J.~Pool, Y.~Kwon, and S.~W. Keckler,
  ``Compressing dma engine: Leveraging activation sparsity for training deep
  neural networks,'' in \emph{HPCA}, 2018.

\bibitem{musae}
B.~Rozemberczki, C.~Allen, and R.~Sarkar, ``{Multi-Scale Attributed Node
  Embedding},'' \emph{Journal of Complex Networks}, 2021.

\bibitem{Sadi2019Efficient}
F.~Sadi, J.~Sweeney, T.~M. Low, J.~C. Hoe, L.~Pileggi, and F.~Franchetti,
  ``Efficient {SpMV} operation for large and highly sparse matrices using
  scalable multi-way merge parallelization,'' in \emph{MICRO}, 2019.

\bibitem{ressparsity}
S.~Salman and X.~Liu, ``Sparsity as the implicit gating mechanism for residual
  blocks,'' in \emph{IJCNN}, 2019.

\bibitem{scalesim}
A.~Samajdar, Y.~Zhu, P.~Whatmough, M.~Mattina, and T.~Krishna, ``{SCALE-Sim:
  Systolic CNN accelerator simulator},'' \emph{arXiv preprint}, 2018.

\bibitem{coraciteseerpubmed}
P.~Sen, G.~Namata, M.~Bilgic, L.~Getoor, B.~Galligher, and T.~Eliassi-Rad,
  ``Collective classification in network data,'' \emph{AI magazine}, 2008.

\bibitem{Silfa2019Neuron}
F.~Silfa, G.~Dot, J.-M. Arnau, and A.~Gonaz\`{a}lez, ``Neuron-level fuzzy
  memoization in {RNNs},'' in \emph{MICRO}, 2019.

\bibitem{Siu2018Memory}
K.~Siu, D.~M. Stuart, M.~Mahmoud, and A.~Moshovos, ``{Memory Requirements for
  Convolutional Neural Network Hardware Accelerators},'' in \emph{IISWC}, 2018.

\bibitem{tensaurus}
N.~Srivastava, H.~Jin, S.~Smith, H.~Rong, D.~Albonesi, and Z.~Zhang,
  ``Tensaurus: A versatile accelerator for mixed sparse-dense tensor
  computations,'' in \emph{HPCA}, 2020.

\bibitem{dblp}
J.~Tang, J.~Zhang, L.~Yao, J.~Li, L.~Zhang, and Z.~Su, ``Arnetminer: Extraction
  and mining of academic social networks,'' in \emph{KDD}, 2008.

\bibitem{cacti}
S.~Thoziyoor, N.~Muralimanohar, J.~H. Ahn, and N.~P. Jouppi, ``{CACTI 5.1},''
  2008-20, HP Labs, Tech. Rep., 2008.

\bibitem{bitmap1}
K.~Ueno, T.~Suzumura, N.~Maruyama, K.~Fujisawa, and S.~Matsuoka, ``Extreme
  scale breadth-first search on supercomputers,'' in \emph{Big Data}, 2016.

\bibitem{graphnlp}
S.~Vashishth, ``Neural graph embedding methods for natural language
  processing,'' \emph{arXiv preprint}, 2019.

\bibitem{wang2021dual}
Y.~Wang, C.~Zhang, Z.~Xie, C.~Guo, Y.~Liu, and J.~Leng, ``Dual-side sparse
  {Tensor Core},'' in \emph{ISCA}, 2021.

\bibitem{overcoming}
X.~Wei, Y.~Liang, and J.~Cong, ``{Overcoming Data Transfer Bottlenecks in
  {FPGA-based DNN} Accelerators via Layer Conscious Memory Management},'' in
  \emph{DAC}, 2019.

\bibitem{ship}
C.-J. Wu, A.~Jaleel, W.~Hasenplaugh, M.~Martonosi, S.~C. Steely~Jr, and
  J.~Emer, ``{SHiP}: Signature-based hit predictor for high performance
  caching,'' in \emph{MICRO}, 2011.

\bibitem{ginconv}
K.~Xu, W.~Hu, J.~Leskovec, and S.~Jegelka, ``How powerful are graph neural
  networks?'' \emph{arXiv preprint}, 2018.

\bibitem{hygcn}
M.~Yan, L.~Deng, X.~Hu, L.~Liang, Y.~Feng, X.~Ye, Z.~Zhang, D.~Fan, and Y.~Xie,
  ``{HyGCN}: A gcn accelerator with hybrid architecture,'' in \emph{HPCA},
  2020.

\bibitem{reddit}
P.~Yanardag and S.~Vishwanathan, ``Deep graph kernels,'' in \emph{KDD}, 2015.

\bibitem{yang2020revisitingoversmoothing}
C.~Yang, R.~Wang, S.~Yao, S.~Liu, and T.~Abdelzaher, ``Revisiting
  over-smoothing in deep gcns,'' \emph{arXiv preprint}, 2020.

\bibitem{scene}
J.~Yang, J.~Lu, S.~Lee, D.~Batra, and D.~Parikh, ``Graph r-cnn for scene graph
  generation,'' in \emph{ECCV}, 2018.

\bibitem{overrated}
X.~Yang, M.~Gao, J.~Pu, A.~Nayak, Q.~Liu, S.~E. Bell, J.~O. Setter, K.~Cao,
  H.~Ha, C.~Kozyrakis, and M.~Horowitz, ``{DNN} dataflow choice is overrated,''
  \emph{arXiv preprint}, 2018.

\bibitem{spzip}
Y.~Yang, J.~Emer, and D.~Sánchez, ``{SpZip}: Architectural support for
  effective data compression in irregular applications,'' in \emph{ISCA}, 2021.

\bibitem{gramer}
P.~Yao, L.~Zheng, Z.~Zeng, Y.~Huang, C.~Gui, X.~Liao, H.~Jin, and J.~Xue, ``A
  locality-aware energy-efficient accelerator for graph mining applications,''
  in \emph{MICRO}, 2020.

\bibitem{snf}
M.~Yoo, J.~Song, H.~Lee, J.~Lee, N.~Kim, Y.~Kim, and J.~Lee,
  ``{Slice-and-Forge: Making Better Use of Caches for Graph Convolutional
  Network Accelerators},'' in \emph{PACT}, 2022.

\bibitem{Yoo2021Making}
M.~Yoo, J.~Song, J.~Lee, N.~Kim, Y.~Kim, and J.~Lee, ``{Making a Better Use of
  Caches for GCN Accelerators with Feature Slicing and Automatic Tile
  Morphing},'' \emph{IEEE CAL}, 2021.

\bibitem{gcod}
H.~You, T.~Geng, Y.~Zhang, A.~Li, and Y.~Lin, ``Gcod: Graph convolutional
  network acceleration via dedicated algorithm and accelerator co-design,''
  \emph{arXiv preprint}, 2021.

\bibitem{graphsaint-iclr20}
H.~Zeng, H.~Zhou, A.~Srivastava, R.~Kannan, and V.~Prasanna, ``{GraphSAINT}:
  Graph sampling based inductive learning method,'' in \emph{ICLR}, 2020.

\bibitem{bitmap2}
J.~Zhang and L.~Gruenwald, ``Regularizing irregularity: Bitmap-based and
  portable sparse matrix multiplication for graph data on gpus,'' in
  \emph{GRADES-NDA}, 2018.

\bibitem{graphp}
M.~Zhang, Y.~Zhuo, C.~Wang, M.~Gao, Y.~Wu, K.~Chen, C.~Kozyrakis, and X.~Qian,
  ``{GraphP: Reducing communication for PIM-based graph processing with
  efficient data partition},'' in \emph{HPCA}, 2018.

\bibitem{edgepred}
M.~Zhang and Y.~Chen, ``Link prediction based on graph neural networks,'' in
  \emph{NeurIPS}, 2018.

\bibitem{Zhang2016CambriconX}
S.~Zhang, Z.~Du, L.~Zhang, H.~Lan, S.~Liu, L.~Li, Q.~Guo, T.~Chen, and Y.~Chen,
  ``{Cambricon-X}: An accelerator for sparse neural networks,'' in
  \emph{MICRO}, 2016.

\bibitem{rrip}
X.~Zhang, C.~Li, H.~Wang, and D.~Wang, ``A cache replacement policy using
  adaptive insertion and re-reference prediction,'' in \emph{ISCA}, 2010.

\bibitem{sparch}
Z.~Zhang, H.~Wang, S.~Han, and W.~J. Dally, ``{SpArch}: Efficient architecture
  for sparse matrix multiplication,'' in \emph{HPCA}, 2020.

\bibitem{Zhou2018CambriconS}
X.~Zhou, Z.~Du, Q.~Guo, S.~Liu, C.~Liu, C.~Wang, X.~Zhou, L.~Li, T.~Chen, and
  Y.~Chen, ``{Cambricon-S}: Addressing irregularity in sparse neural networks
  through a cooperative software/hardware approach,'' in \emph{MICRO}, 2018.

\bibitem{Zhu2019Sparse}
M.~Zhu, T.~Zhang, Z.~Gu, and Y.~Xie, ``{Sparse Tensor Core}: Algorithm and
  hardware co-design for vector-wise sparse neural networks on modern {GPUs},''
  in \emph{MICRO}, 2019.

\bibitem{gridgraph}
X.~Zhu, W.~Han, and W.~Chen, ``{GridGraph}: Large-scale graph processing on a
  single machine using 2-level hierarchical partitioning,'' in \emph{USENIX
  ATC}, 2015.

\bibitem{graphq}
Y.~Zhuo, C.~Wang, M.~Zhang, R.~Wang, D.~Niu, Y.~Wang, and X.~Qian, ``{GraphQ:
  Scalable PIM-based graph processing},'' in \emph{MICRO}, 2019.

\end{thebibliography}
% \setlength{\footskip}{700pt} % don't know why -- but \balance was causing the page number (13) off the position
%%%%%%%%%%%%%%%%%%%%%%%%%%%%%%%%%%%%

\end{document}